\documentclass{article}
\usepackage[utf8]{inputenc}
\usepackage{amsmath}
\usepackage{amsfonts}
\usepackage{amssymb}
\usepackage{graphicx}
\usepackage[ruled,vlined]{algorithm2e}
\SetKwRepeat{Do}{do}{while}

\title{A global-local neighborhood search algorithm and tabu search for flexible job shop scheduling problem}
\author{Juan Carlos Seck-Tuoh-Mora*, Nayeli J. Escamilla-Serna, \\
Joselito Medina-Marin, Norberto Hernandez-Romero, \\
Irving Barragan-Vite, Jose R. Corona-Armenta\\
AAI-ICBI-UAEH. Carr Pachuca-Tulancingo Km 4.5. \\
Pachuca 42184 Hidalgo. Mexico\\}
\date{October 2020}

\begin{document}

\maketitle

\begin{abstract}
The Flexible Job Shop Scheduling Problem (FJSP) is a combinatorial problem that continues to be studied extensively due to its practical implications in manufacturing systems and emerging new variants, in order to model and optimize more complex situations that reflect the current needs of the industry better. This work presents a new meta-heuristic algorithm called GLNSA (Global-local neighborhood search algorithm), in which the neighborhood concepts of a cellular automaton are used, so that a set of leading solutions called $smart\_cells$ generates and shares information that helps to optimize instances of FJSP. The GLNSA algorithm is complemented with a tabu search that implements a simplified version of the Nopt1 neighborhood defined in \cite{mastrolilli2000effective} to complement the optimization task. The experiments carried out show a satisfactory performance of the proposed algorithm, compared with other results published in recent algorithms and widely cited in the specialized bibliography, using $ 86 $ test problems, improving the optimal result reported in previous works in two of them.
\end{abstract}


Submitted to: PeerJ Computer Science



\section{Introduction}


The scheduling of jobs and resource assignments in a production system includes a series of combinatorial problems that continue to be widely investigated today to propose and test new meta-heuristic algorithms.

One of these problems is the Flexible Job Shop Scheduling Problem (FJSP), an extension of the Job Shop Scheduling Problem. This problem consists of assigning a set of jobs to be processed on multiple machines. Each job consists of several operations that must be processed sequentially. Operations of different jobs can be interlarded in the scheduling. The system's flexibility is given with the possibility that each operation (perhaps all of them) of a given set can be processed in several machines.

Thus, the FJSP aims to find the best possible machine assignment and the best possible scheduling of operations; classically, the objective is to find the shortest possible time (or makespan) to process all the jobs.


The FJSP continues to be a very active subject of research, since its first definition in \cite{brucker1990job}. Many of the latest works have focused on presenting hybrid techniques to find better results. For example, an algorithm has been proposed in which genetic operators and the tabu search \cite{li2016effective} interact. Another work combines discrete bee colony operators with tabu search to optimize classic problems and problems with job cancellation and machine breakdowns \cite{li2017hybrid}. Other recent work uses discrete particle swarm operators with hill climbing and random restart to optimize well-known test problems \cite{kato2018new}. These references are just a small sample of the works that develop hybrid techniques of discrete operators and local search to propose new algorithms that improve the makespan's calculation in FJSP instances.


Within the meta-heuristic algorithms, a new optimization strategy has been proposed using concepts of cellular automata. Relevant work is proposed in \cite{shi2011cellular}, where different types of cellular automaton-like neighborhoods are used in conjunction with particle swarm operations for continuous global optimization.

Other related work using cellular automaton-like neighborhoods to design of type IIR filters can be consulted in \cite{lagos2017new}. Recently, another discrete optimization algorithm has been presented for the concurrent layout scheduling problem in the job-shop environment \cite{hernandez2020solution}, where the idea of cellular automata inspires the neighborhood strategy used by the proposed algorithm.

The idea that a solution can be improved by generating a neighborhood with new solutions which are generated by small changes in its current information and by sharing information with other solutions is the inspiration behind these algorithms, like the evolution rule of a cellular automaton \cite{mcintosh2009one}.


Following this trend developed in previous studies, this work proposes applying this neighborhood idea to optimize instances of the FJSP. Specifically, an algorithm is proposed that uses a set of leading solutions called $smart\ _cells$. In each iteration of the algorithm, the population of $smart\_cells$ is selected using elitism and tournament.

With this selected population, each $smart\_cell$ generates a neighborhood of new solutions using classical operators of combinatorial problems (insertion, exchange, and path-relinking). The best one is selected from this neighborhood, which updates the $smart\_cell$ value.

The neighborhood-based optimization of each $smart\_cell$ is complemented by a tabu search using a simplified version of the Nopt1 neighborhood proposed in \cite{mastrolilli2000effective}. In this neighborhood, a random critical path is selected, and a better solution is searched for, perhaps a machine minimizing the makespan for each operation on the critical path but without changing its position in the current array of operations.

This neighborhood management of each $smart\_cell$ and the simplified neighborhood Nopt1 allows us to obtain an algorithm of less complexity than those previously proposed, which can adequately solve the test problems commonly used in the specialized literature. In particular, for instances of FJSP with high flexibility (where more machines can perform the same operation), two solutions are presented with better makespan values compared to the algorithms reported in this manuscript.

The structure of the paper is as follows. Section \ref{secc:Estado_arte} presents a state of the art of the FJSP. Section \ref{secc:Formulación} describes the FJSP formulation. Section \ref{secc:GLNSA} explains the strategy and operators used for global and local searches that define the GLNSA. Section \ref{secc:Experiment} shows the experimental results obtained when GLNSA is applied for instances of the FJSP with high flexibility. The last section provides the conclusions of the article and prospects for future work.

\section{State of the art of FJSP}
\label{secc:Estado_arte}

The FJSP is an extension of the classic Job-shop Scheduling Problem (JSP). The classic problem seeks to find the assignment of operations in a set of predefined machines, while the flexible case consists of a sequence of operations, where each operation can be performed on several available machines, possibly with different processing times. To solve an instance of the FJSP, one must consider two sub-problems, assignment and scheduling \cite{brandimarte1993routing}. For each operation, the first assigns a machine from a set of available machines. The second is in charge of sequencing the operations assigned to each machine to obtain a feasible schedule to minimize the objective function \cite{li2014discrete}.

The problem definition was introduced by \cite {brucker1990job}, who proposed a polynomial-graphical algorithm to solve a problem with only two jobs, concluding that FJSP belongs to the category of NP-hard problems for which there are no algorithms that can bring optimal solutions in polynomial time.


One of the first works to address the FJSP problem with a heuristic approach is \cite{brandimarte1993routing}, which uses dispatch rules and a hierarchical tabu search algorithm to solve the problem and introduce $15$ instances.

Since then, many investigations have addressed the FJSP problem and applied different approaches and methods to solve it. For example, \cite{mastrolilli2000effective} introduces two neighborhood functions to use local search techniques by proposing a tabu search procedure (TS).

A practical hierarchical solution approach is proposed in \cite{xia2005effective} to solve multiple targets for the FJSP problem. The proposed approach uses particle swarm optimization (PSO) to assign operations in machines and the simulated annealing (SA) algorithm to each machine's program operations. The objective is to minimize the makespan (maximum completion time), the total machine workload, and the critical machine workload.

A genetic algorithm (GA) is proposed in \cite{pezzella2008genetic} applied to the FJSP. The developed algorithms integrate different selection and reproduction strategies and show that an efficient algorithm is developed when combining different rules to find the initial population, selection, and reproduction operators.

Another hybridized genetic algorithm is described in \cite{gao2008hybrid}, which strengthens the search for individuals and is improved with the variable neighborhood descent variable (VND); since it is a multi-objective problem, they seek the minimum makespan, maximum workload and minimum total workload. Two local search procedures are used, the first for a moving operation and the second for two moving operations.

Hybridization of two algorithms, PSO and TS, are combined in \cite{zhang2009effective} to solve a multi-objective problem, that is, several conflicting objectives, mainly in large-scale problems, where the PSO has a high search efficiency combining local search and global search and TS is used to find a near-optimal solution.

In \cite{amiri2010variable}, a variable neighborhood search algorithm (VNS) applied to the FJSP is proposed, and its objective function is the makespan. Several types of neighborhoods are presented, where they use assignment and sequence problems to generate neighboring solutions.

Another hybrid algorithm using TS and VNS is presented in \cite{li2010effective}, which considers three minimization objectives, produces neighboring solutions in the machine assignment module, and performs local searches in the operation scheduling.

An algorithm that treats the FJSP considering parallel machines and maintenance costs is exposed in \cite{dalfard2012two}, which proposes a new mathematical model that applies the hybrid genetic algorithm (HGA) and the SA algorithm, obtaining satisfactory results in $12$ experiments using multiple jobs.

The work in \cite{yuan2013hybrid} adapts the harmony search algorithm (HS) in the FJSP problem. They developed techniques to convert the continuous harmony vector into two vectors, and these vectors are decoded to reduce the search space applied to an FJSP problem. Finally, they introduce an initialization scheme by combining heuristic and random techniques and incorporating the local search in the HS, in order to speed up the local search process in the neighborhood.

Another discrete algorithm based on an artificial-bee colony, called DABC, is presented in \cite{li2014discrete}. They take three objectives as criteria, where they adopt a self-adaptive strategy, represented by two discrete vectors and a TS, demonstrating that its algorithm is efficient and effective with high performance.

A genetic algorithm that incorporates the Taguchi method in its coding to increase its effectiveness is exposed in \cite{chang2015solving} and it evaluates the performance of the proposed algorithm using the results of \cite{brandimarte1993routing}.

A hybrid evolutionary algorithm based on the PSO and the Bayesian optimization algorithm (BOA) is developed in \cite{sun2015bayesian} and used to know the relationship between the variables and its objective to minimize the processing time and improve the solutions and the robustness of the process.

A multi-objective methodology is described in \cite{ahmadi2016multi} for the FJSP programming problem in the specific case of machine breakdown situations. They use two algorithms, the non-dominated sorting genetic algorithm (NSGA) and the NGSA-II, which is usually utilized to solve large multi-objective problems, like evaluating the status and condition of machine breakdowns.

A hybrid algorithm that uses a genetic algorithm and tabu search to minimize the makespan is presented in \cite{li2016effective}. The proposed algorithm has adequate search capacity and balances intensification and diversification very well.

In \cite{li2017hybrid}, the HABC algorithm, a hybrid artificial bee colony (ABC) algorithm and the improved TS algorithm are proposed to solve the FJSP in a textile machine company. Three rescheduling strategies are introduced, schedule reassembly, schedule intersection, and schedule insertion, to address dynamic events such as new jobs inserted, old jobs, and to address when there may be cell and machine breakdowns. The HABC algorithm is shown to have satisfactory exploitation, exploration, and performance to solve the FJSP.

A non-dominant genetic classification algorithm that serves as an evolutionary guide for an artificial bee colony (BEG-NSGA-II) is developed in \cite{deng2017bee}, and it focuses on the multi-objective problem (MO-FJSP). Usually, this type of algorithm converges prematurely to the local solution. Therefore, that paper uses a two-stage optimization to avoid these disadvantages in order to minimize the maximum completion time, the workload of the most loaded machine, and the total workload of all the machines.

An optimization algorithm applying a hybrid ant colony optimization (ACO)  to solve the FJSP described in \cite{wu2017flexible} is based on a 3D disjunctive graph, and has four objectives: to minimize the completion time, the delay or anticipation penalty cost, average machine downtime, and the cost of production.

In \cite{shen2018solving} the FJSP is addressed using sequence-dependent setup time (SDST) and a Mixed Integer Linear Programming model (MILP) to minimize the makespan using the TS as an optimization algorithm. They apply specific functions and a diversification structure, comparing their model with well-known reference instances and two meta-heuristics from the literature, obtaining satisfactory results.

In \cite{kato2018new}, new strategies are used in population initialization, particle displacement, stochastic assignment of operations, and partially and fully flexible scenario management, to implement a hybrid algorithm using the PSO for the machine routing subproblem and explore the solution space with a Random Restart Hill Climbing (RRHC) for the local search programming subproblem.

A new definition of the FJSP (double flexible job-shop scheduling problem, DFJSP) is described in \cite{gong2018new}. Here, the processing time is considered, and factors related to the environment's protection are presented as an indicator. They presented and resolved ten benchmarks using the new algorithm.

An algorithm that combines the uncertainty processing time to solve an FJSP in order to minimize uncertain times and the makespan is presented in \cite{xie2018flexible}, the algorithm uses gray information based on external memory with elitism strategy.

In \cite {reddy2018effective} the FJSP is solved for the minimization of the makespan and the workload of the machines using a programming model (mixed-integer non-linear programming, MINLP) with machines focused on real-time situations using a new hybrid algorithm through PSO and GA to solve multiple objectives, obtaining high-quality solutions.

An algorithm that addresses the FJSP to minimize the total workflow and inventory costs is described in \cite{meng2018hybrid}; it applies an artificial bee colony (ABC) and the modified migratory bird algorithm (MMBO), to obtain a satisfactory capacity search.

In \cite{tang2019flexible}, two optimization methods are applied with two significant characteristics in practical casting production, the Tolerated Time interval (TTI) and the Limited Start Time interval (LimSTI).

A model to calculate the energy consumption of machinery is presented in \cite{wu2019flexible}, which has different states and deterioration effect to determine the real processing time to apply a hybrid optimization using the SA algorithm.

A new algorithm called hybrid multi-verse optimization (HMVO) is proposed in \cite{lin2019hybrid} to treat a fuzzy problem in an FJSP. Route linking technique is used, and a mixed push-based phase to expand the search space and local search to improve the solution is incorporated.

Another elitist non-dominated sorting hybrid algorithm (ENSHA) for a multi-objective FJSP problem is explained in \cite{li2019elitist}. A configuration dependent on the sequence is used. Its objective is to find the minimum makespan and the total costs of the installation, by proposing a learning strategy based on estimating the distribution algorithm (EDA) and checking its effectiveness with $39$ instances and a real case study.

Previous works show that the algorithms dedicated to solving FJSP instances and have had the best results use hybrid techniques that combine meta-heuristic techniques. Another point analyzed in the literature review is that the most recurrent objective function is the makespan as the most widely used performance measure.

\section{Problem formulation}
\label{secc:Formulación}

The FJSP is presented following the definition of \cite{zuo2017adaptive}. There is a set of $n$ jobs $J = \{J_1, J_2, \ldots J_n\}$ and a set of $m$ machines $M=\{M_1, M_2, \ldots M_m \}$. Each $J_i$ job consists of a sequence of operations $O_ {J_i} = \{O_{i, 1}, O_{i, 2}, \ldots, O_{i, n_i} \}$ where $n_i $ is the number of operations contemplated by the job $J_i$. For $1 \leq i \leq n$ and $1 \leq j \leq n_i $, each operation $O_{i, j}$ can be processed by one machine from a set of machines $M_{i, j} \subseteq M$. The processing time of $O_{i, j} $ on the machine $M_k$ is denoted by $p_ {i, j, k}$.

In an instance of the FJSP, the following conditions are considered:
\begin{enumerate}
\item An operation cannot be interrupted while a machine is processing it. \item One machine can process at most one operation.
\item Once the order of operations has been determined, it cannot be modified.
\item Breakdowns in machines are not considered
\item The works are independent of one another.
\item The machines are independent of one another.
\item The time used for the machines' preparation and the transfer of operations between them is negligible.
\end{enumerate}

A solution of the FJSP is defined as the order of operations $O_{i,j}$ that respects each job's precedence restrictions. For each operation, $ O_{i,j}$ a machine is selected from the subset $M_ {i,j}$. The objective is to find the feasible order of operations $O_ {i,j}$ and for each operation, the assignment of a machine in $M_{i,j}$ minimizing the makespan, or the time needed to complete all jobs. The makespan can be formally defined as $C_{max} = max\{C_i \} $ where $C_i$ represents the completion time for all operations of the job $J_i$, for $ 1 \leq i \leq n $.

Thus, an instance of the FJSP involves two problems, the scheduling of operations and the machine assignment to each operation.

A recent strategy for solving this type of problem is the hybridization of techniques that optimize both problems. This manuscript follows this research line, proposing an algorithm that combines the generation of new solutions using a neighborhood inspired by cellular automata, complemented with a local search technique based on tabu search using a simplified version of the Nopt1 neighborhood presented in \cite{mastrolilli2000effective}.

This type of neighborhood allows the exploration of new solutions using well-known operators for the scheduling problem. The conjunction with the simplified neighborhood Nopt1 allows proposing a neighborhood-based algorithm that performs global and local searches in each iteration, with a complexity similar to the most recent algorithms, and uses less computational time, obtaining satisfactory results for problems with high flexibility.

\section{Global-local neighborhood search algorithm for the FJSP}
\label{secc:GLNSA}


\subsection{General description of the GLNSA}

The proposed algorithm is based on the neighborhood concept in cellular automata similar to \cite{hernandez2020solution}. A cellular automaton is a discrete dynamic system made up of indivisible elements called cells, where each of them changes its state over time. The state change can depend on both the current state of each cell and its neighboring cells. With such simple dynamics, cellular automata are capable of creating periodic, chaotic, or complex global behavior \cite{wolfram2002new} \cite{mcintosh2009one}.

\begin{figure}[htbp]
\begin{center}
\includegraphics[scale=0.44]{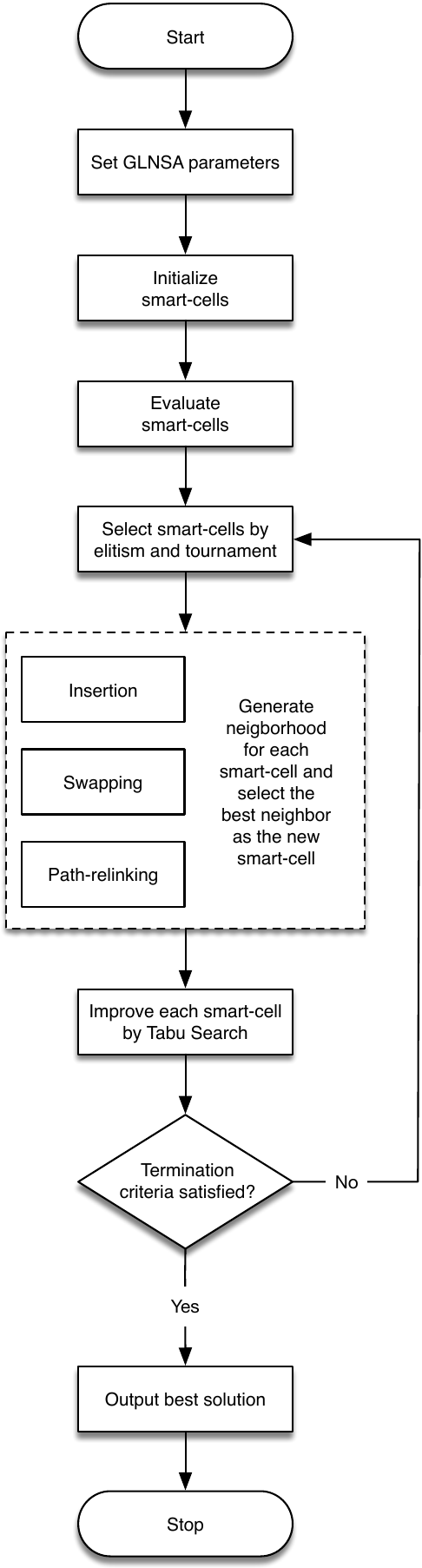}
\caption{The workflow of the GLNSA.}
\label{fig:GeneralFlowChart}
\end{center}
\end{figure}

In this work, the idea of a cellular-automaton neighborhood is an inspiration to propose a new algorithm that optimizes instances of the FJSP. The algorithm has $S_n$ leading solutions or $smart\_cells$, where each of them first performs a global search mainly focused on making modifications to the sequencing of operations, applying several operators to form a neighborhood, and selecting the best modification. Then, each $smart\_cell$ executes a local search focused on machine assignment, applying a tabu search with increasing iterations on each step of the algorithm. Thus, in each iteration, a global exploration search and a local exploitation search are performed to optimize instances of the FJSP, and hence, this process is called the Global-Local Neighborhood Search Algorithm (GLNSA). The corresponding flow chart is described in Figure \ref{fig:GeneralFlowChart}.

The detailed explanation of the encoding and decoding of solutions, the neighborhood used, its operators for the global search, and the tabu search operators are presented in the following sub-sections. The general procedure of GLNSA is in Algorithm \ref{alg:GLNSA}.

\begin{algorithm}[H]
\SetAlgoLined
\KwResult{Best $smart\_cell$}
 Set the parameters of the GLNSA\;
 Initialize the population of $smart\_cell$s with $S_n$ solutions generated at random\;
 Evaluate each $smart\_cell$ to obtain its makespan\;
 
 \Do{(Iteration number less than $G_n$ or stagnation number less than $S_b$)}
 {
 Select a refined population from the best $smart\_cells$ using elitism and tournament \;
 For each $smart\_cell$, generate a neighborhood (using insertion, swapping, and path relinking operators) and take the best neighbor as new $smart\_cell$\;
 For each $smart\_cell$, improve the machine assignment by tabu search\;
 }
 Return the $smart\_cell$ with minimum makespan \;
\caption{General description of the GLNSA}
\label{alg:GLNSA}
\end{algorithm}

\subsection{Encoding and decoding of solutions}

To represent a solution of an instance of the FJSP, we take the encoding with two strings ($OS$ and $MS$) described in \cite{li2016effective},  $OS$ for operations and $MS$ for machines.

The string $OS$ consists of a permutation with repetitions where each job $J_i $ appears $n_i$ times. The string $OS$ is read from left to right, and the $j-th$ appearance of $J_i$ indicates that the $O_ {i,j}$ operation of job $J_i$ should be processed. This coding of the sequence of operations $OS$ has the advantage that any permutation with repetitions produces a valid sequence so that the operators used in this work will always yield a feasible sequence.

The string $MS$ consists of a string with a length equal to the number of the total operations. The string is divided into $n$ parts, where the $i-th$ part contains the machines assigned to $J_i$ with $n_i$ elements. For each $i-th$ part of $MS$, the $j-th$ element indicates the machine assigned to $O_{i,j}$.


Initially, each solution $OS$ is generated at random, making sure that each job $J_i$ appears exactly $n_i$ times. Once $OS$ is created, the sequence is read from left to right; the $j-th$ appearance of $J_i$ corresponds to the operation $O_{i,j}$. For every operation, one of the possible machines that can process it is selected at random. That machine is assigned to the element $j$ of the $i-th$ part in $MS$.

\begin{figure}[htbp]
\begin{center}
\includegraphics[scale=0.27]{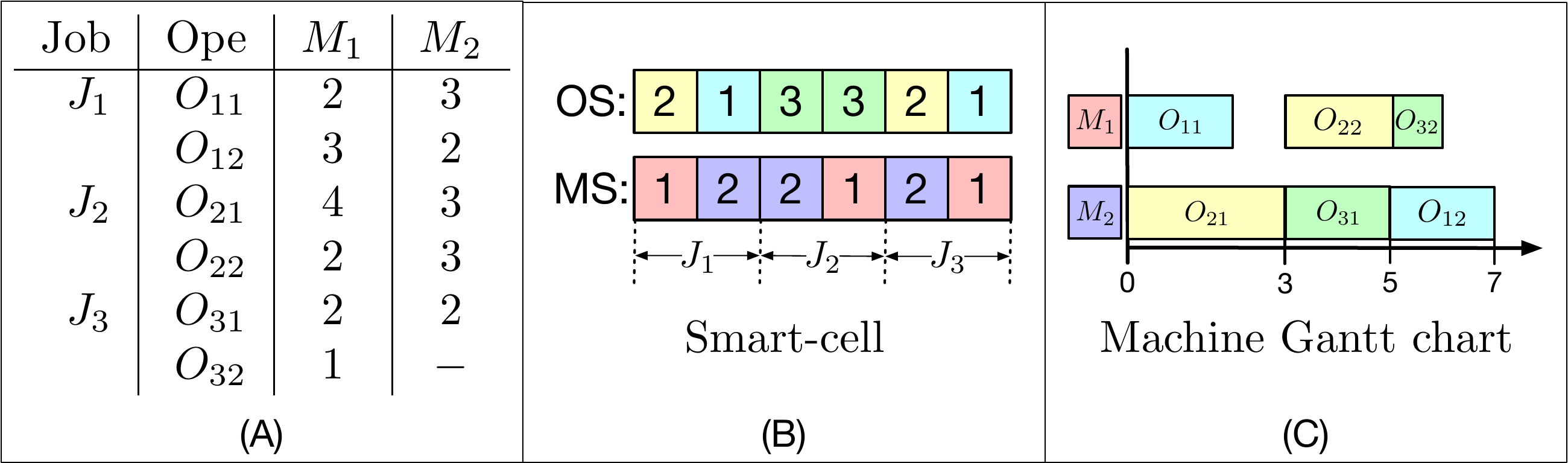}
\caption{Example of encoding and active decoding used by the GLNSA.}
\label{fig:ExampleActiveScheduling}
\end{center}
\end{figure}

For each pair of $OS$ and $MS$ strings that represents a solution for an instance of the FJSP, their decoding is done using an active scheduling. For each operation $O_{i,j}$ in $OS$ and its assigned machine $k$ in $MS$, its initial time $s(O_ {i,j})$ is taken as the greater time between the completion time  of the previous operation $O_{i,j-1}$ and the lesser time available in the machine $k$ (not necessarily after the last operation programmed in that machine) where there is an available time slot, such that the processing time $p_{i,j,k}$ is less than or equal to the size of this slot. The time the $O_{i,j}$ operation is completed is called $C(O_{ij})$, and for $j=1$, $s(O_ {i,j-1})=0$ for all $1 \leq i \leq n$. An example with $3$ jobs and $2$ machines is shown in Fig. \ref{fig:ExampleActiveScheduling}.

In Fig. \ref{fig:ExampleActiveScheduling} (A), there are $3$ jobs, each with $2$ operations. Almost every operation can be executed on the $2$ available machines. On average, each trade can be executed for $ 1.83 $ machines; this is called the system's flexibility. Note that the time at which each operation is performed on each machine may be different. In part (B), you can see how a $ smart\_cell $ is encoded. It consists of two strings; the first $ OS $ is a permutation with repetitions, where each job appears $ 2 $ times. The second string $ MS $ contains the machines programmed for each operation, where the first $ 2 $ elements correspond to the machines assigned to the operations $ O_ {1,1} $ and $ O_ {1,2} $, the second block of $ 2 $ elements specify the machines assigned to operations $ O_ {2,1} $ and $ O_ {2,} $, and so on. Finally, part (C) indicates the decoding of the $ smart\_cell $ reading of the string $ OS $ from left to right. In this case, the $ O_{22} $ operation, which is the fifth task programmed in $ OS $, is actively accommodated, since there is a gap in the $ M_1 $ machine between $ O_{11} $ and $ O_{32} $, which is of sufficient length to place $ O_{22} $ just after the preceding operation on the machine $ M_2 $ and without moving the operations already programmed in $ M_1 $. This scheduling gives a final makespan of $ 7 $ using the active decoding.

\subsection{Selection method}

The GLNSA uses elitism and tournament to refine the population, by considering the value of the makespan of each $ smart\_cell $. For elitism, a proportion $ E_p $ of $ smart\_cells $ is taken with the best values of population's makespan. Those solutions will remain unchanged in the next generation of the algorithm. The rest of the population members are selected using a tournament scheme, where a group of $ b $ $smart\_cells$ are randomly selected from the current population and competed with each other, and the $smart\_cell $ with the best makespan is selected to become part of the population to be improved using the global and local search operators described later. In this work, we take $ b = 2 $. This mixture of elitism and tournament allows a balance between exploring and exploiting the information in the population, keeping the best $smart\_cells $, and allowing the other $ smart\ _cells $ with a good makespan to continue in the optimization process.

\subsection{Neighborhood structure}

In this section, two neighborhood structures are presented for the FJSP. The first neighborhood focuses mainly on the sequencing of operations and random assignments of machines to each operation, and the second neighborhood focuses on a local search neighborhood to improve the assignment of the machines once the sequence of operations has been modified.

For the second neighborhood, a simplification of the $ Nopt1 $ neighborhood and the makespan estimation explained in \cite{mastrolilli2000effective} are used to improve the GLNSA execution time.


\subsection{Exploration neighborhood}

\begin{figure}[htbp]
\begin{center}
\includegraphics[scale=0.25]{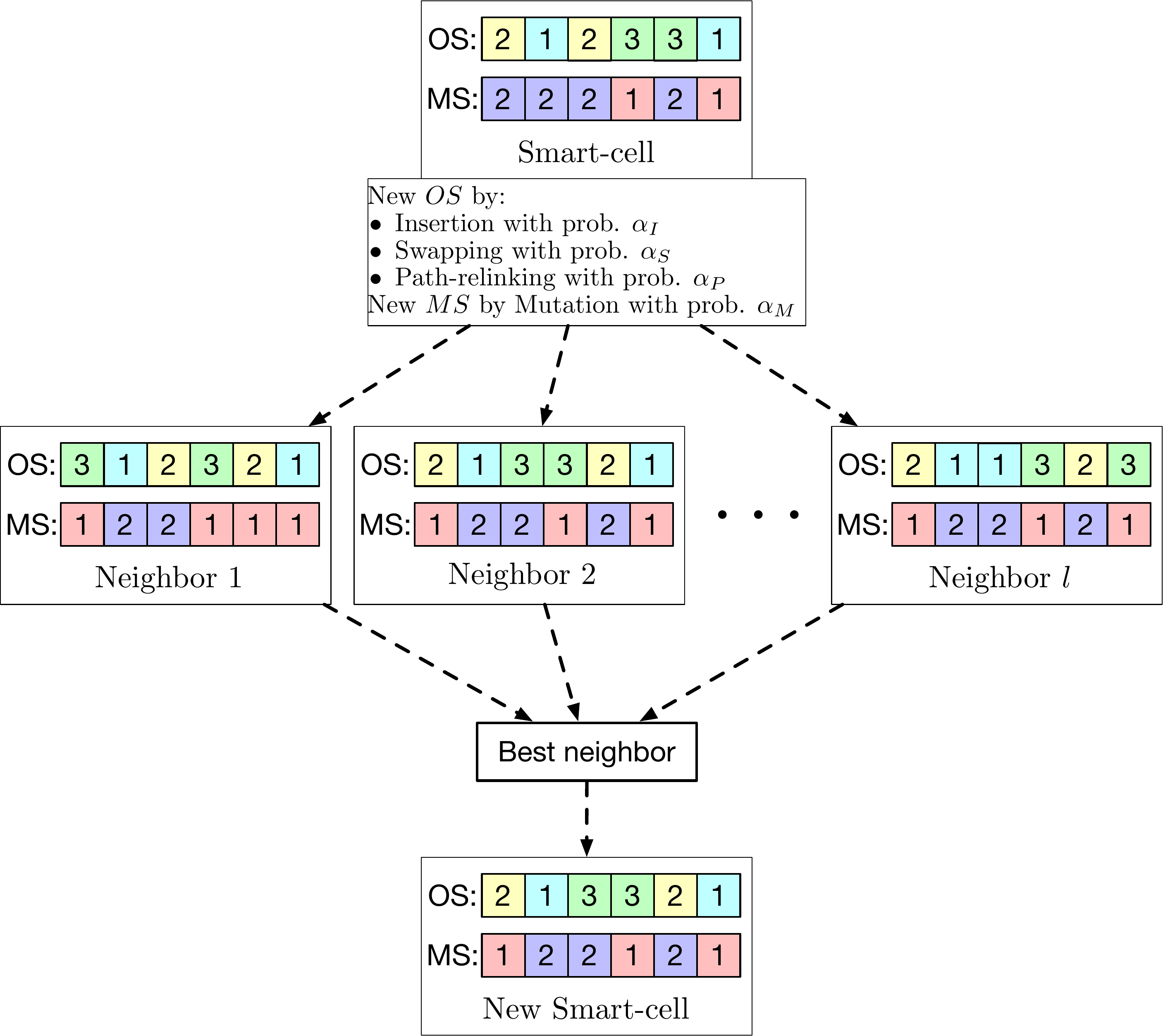}
\caption{Exploration neighborhood used by the GLNSA.}
\label{fig:VecindadExploracion}
\end{center}
\end{figure}

For the global search neighborhood, well-known operators used in various task sequencing, inserting, swapping, and path-relinking problems are used. These operators are used to optimize the sequence of operations. For the machine assignment, the mutation operator used in \cite{li2016effective} is applied. Each $smart\_cell$ generates $l$ neighbors using one of the three possible operators (insertion, swapping or path-relinking) with probability $ \alpha_I $, $ \alpha_S $ and $ \alpha_P $ respectively to generate a variant of $OS$. For the machine assignment, the mutation operator with probability $ \alpha_M $ is employed to generate another variant of $ MS $. From these $ l $ neighbors, the one with the smallest makespan is chosen, and this will be the new $smart\_cell $. This neighborhood is exemplified in Fig. \ref{fig:VecindadExploracion}.

\subsection{Insertion}

\begin{figure}[htbp]
\begin{center}
\includegraphics[scale=0.3]{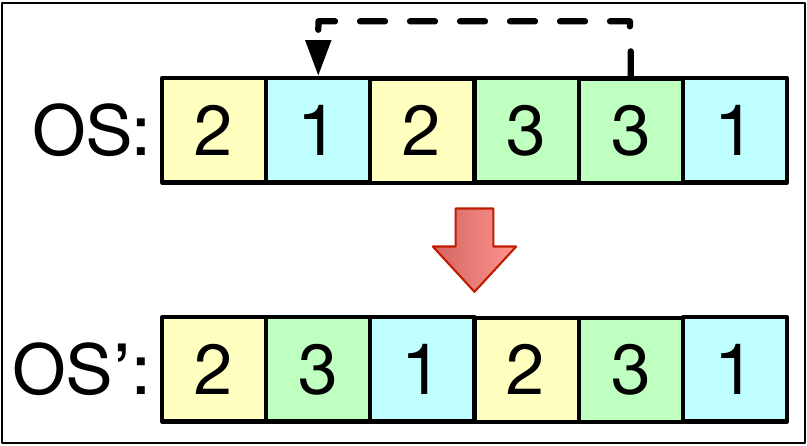}
\caption{The insertion operator.}
\label{fig:Insercion}
\end{center}
\end{figure}

The insertion operator consists of selecting $ 2 $ different positions $ k_1 $ and $ k_2 $ of the string $ OS $ in a $smart\_cell $ to obtain another string $ OS'$. For example, if $ k_1> k_2 $ with $ OS = (O_1, \ldots O_{k_2} \ldots O_ {k_1} \ldots) $, then we get the string $ OS'=(O_1, \ldots O_{k_1 } O_{k_2} \ldots O_{k_1-1} \ldots) $. This is analogous if $ k_1 < k_2 $. This is exemplified in Fig. \ref{fig:Insercion}.

\subsection{Swapping}

\begin{figure}[htbp]
\begin{center}
\includegraphics[scale=0.3]{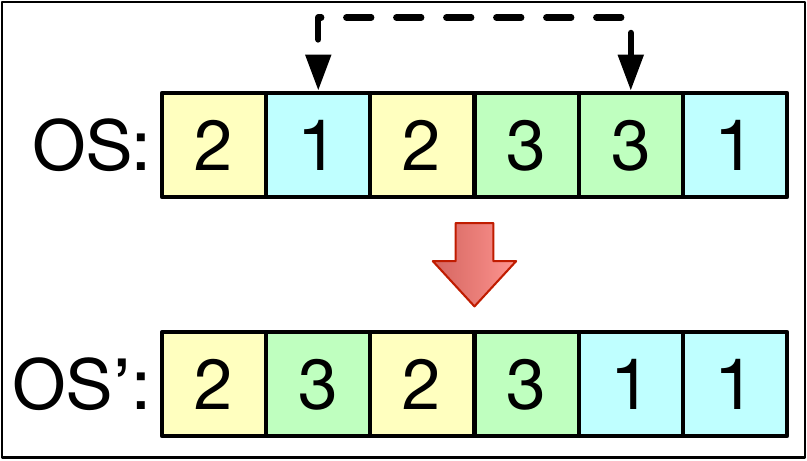}
\caption{The swapping operator.}
\label{fig:Intercambio}
\end{center}
\end{figure}

The swapping operator consists of selecting $ 2 $ different positions $ k_1 $ and $ k_2 $ from the string $OS$ to exchange their positions. For example, if $ k_1 <k_2 $ with $ OS = (O_1, \ldots O_{k_1} \ldots O_{k_2} \ldots) $, then the string $ OS'= (O_1, \ldots O_{k_2 } \ldots O_{k_1} \ldots) $. This is analogous if $ k_1> k_2 $. This is exemplified in Fig. \ref{fig:Intercambio}.

Insertion and swapping are operators classically used in meta-heuristics for task scheduling problems, since they provide good quality solutions \cite{blazewicz1996job} \cite{cheng1999tutorial} \cite{deroussi:hal-00678053}.

\subsection{Path relinking}

Path relinking (PR) involves establishing a route between two different $ smart\_cells $ for their strings $ OS $ and $ OS '$. The route defines intermediate strings ranging from $ OS $ to $ OS' $. To form this route, the first position $ k $ of $ OS $ is taken such that $ OS_k \neq OS'_k $. Then the first position $ p $ is located after $ k $ such that $ OS_p = OS'_k $. Once both positions have been found, the positions $ k $ and $ p $ are exchanged in $ OS $ to obtain a new string closer to $ OS '$. The PR is repeated until $ OS '$ is obtained. In the end, one of the intermediate solutions is randomly chosen.

\begin{figure}[htbp]
\begin{center}
\includegraphics[scale=0.23]{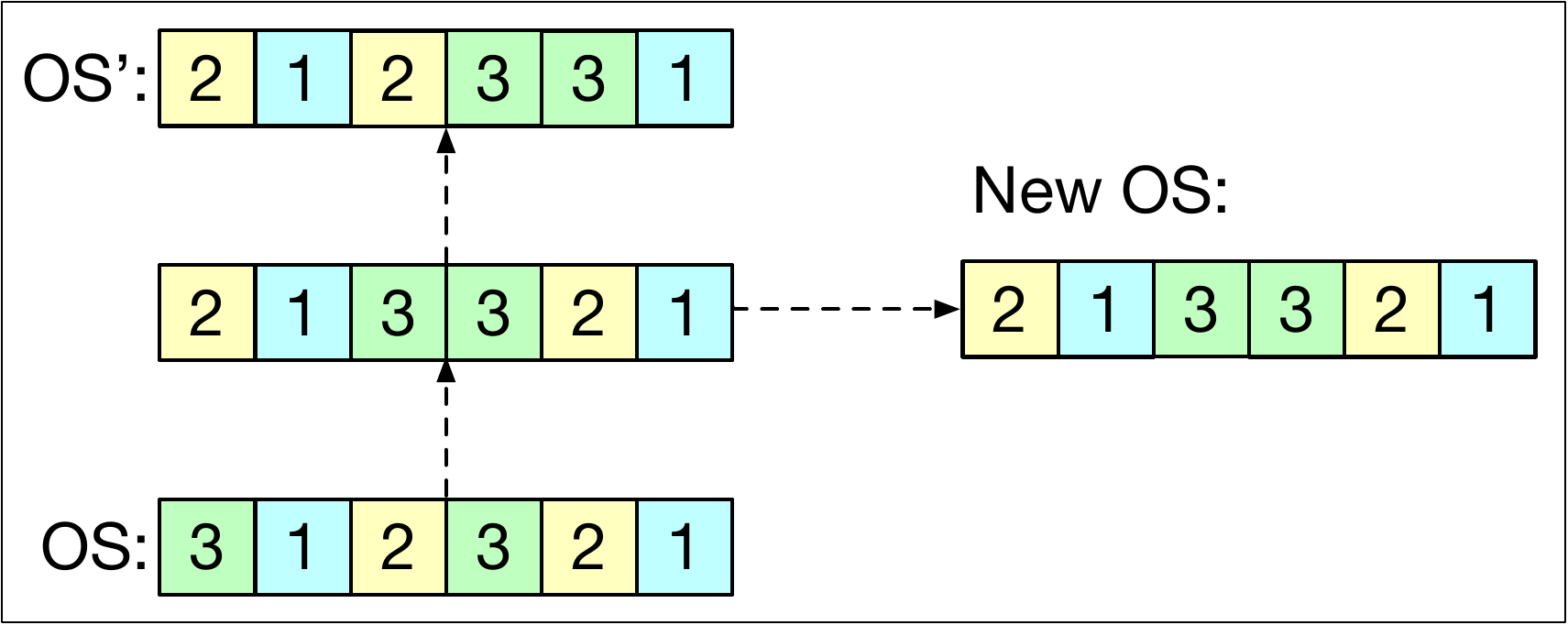}
\caption{The path relinking operator.}
\label{fig:Path_relinking}
\end{center}
\end{figure}

The idea of the PR is to generate solutions that combine the information of $ OS $ and $ OS '$ and it fulfills two objectives in the GLNSA: if both $ smart\_cells $ have similar machine strings, PR acts as a local search method that refines the strings of operations. On the other hand, if both $ smart\_cells $ have very different machine strings, PR works as an exploration method that generates new variants of one of the $ smart\_cell $, taking the other as a guide, as shown in the Fig. \ref{fig:Path_relinking}. PR has already been used successfully in the FJSP, as shown in its application with different neighborhood variants in \cite{gonzalez2015scatter}.

\subsection{Mutation of the string of machines}

For the $ MS $ string of each $smart\_cell$, a mutation operator is applied that selects half of the positions in $MS$ at random. For each position linked to an operation in $OS$, the assigned machine is changed for another selected at random so that it can perform the respective operation in $OS$.

\begin{figure}[htbp]
\begin{center}
\includegraphics[scale=0.26]{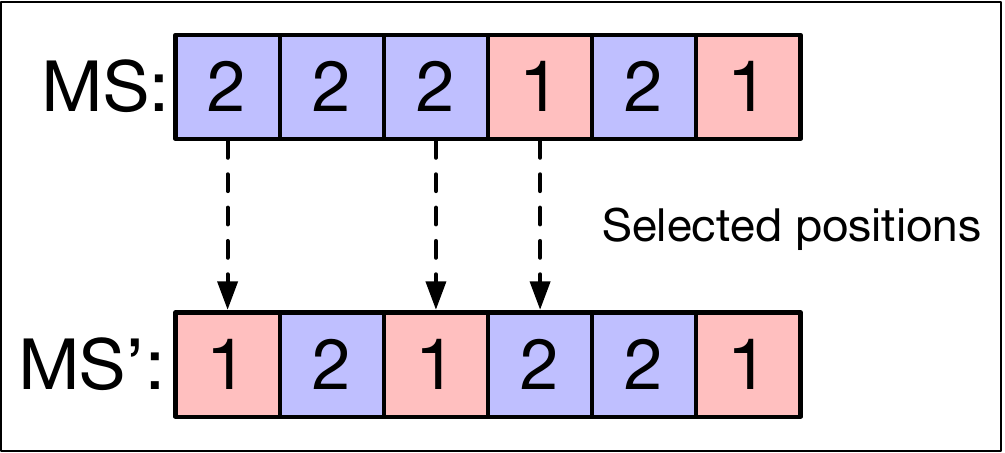}
\caption{Mutation operator in machine strings.}
\label{fig:Mutacion}
\end{center}
\end{figure}

Figure \ref{fig:Mutacion} shows an example of the mutation operator applied to a sequence $MS$. Three machines are selected corresponding to the operations $O_{11}$, $O_{21}$ and $O_{22}$ in $OS$. The new machine assignment is obtained by changing the machines in these positions to others that can also perform these operations.

\subsection{Local search neighborhood}

For the task of exploiting the $smart\_cells$ information, the tabu search scheme (TS) \cite{glover1998tabu} is used, which is widely applied for combinatorial problems given its simplicity and efficiency. TS allows generating solutions that may not improve the makespan of the original $smart\_cell$ as long as the operation and the machine selected to obtain the new solution are not forbidden. TS must keep a record of the operation and the machine selected in each movement and the threshold at which this movement will remain tabu, and the aspiration criterion that allows a solution to be accepted even if it is tabu.

In this work, the implementation of TS is resumed using a simplification of the neighborhood structure proposed in \cite{mastrolilli2000effective} and the makespan estimation explained in the same work, to reduce the computational time of TS. The general TS procedure is described in the algorithm \ref{alg:TS}.

\begin{algorithm}[H]
\SetAlgoLined
\KwResult{Best $smart\_cell$}
 Take a $smart\_cell$ as initial solution\;
 Take the operations and their machine assignments from a critical path of the $smart\_cell$\;
 Initialize TS (empty list of operations/machines, iteration number $T_n$ for the TS and tabu threshold $T_u$)\;
 Set $new\_solution=smart\_cell$ \;
 \For{$T_n$ iterations}
 {
 Generate new neighborhoods of $new\_solution$ modifying only the assigned machines to the critical operations\;
 Select the $best\_neighbor$\;
 \eIf{$best\_neighbor$ holds the aspiration criterion or is not tabu}
 {
 Set $best\_neighbor$ as $new\_solution$\;
 }
 {
 Select the oldest tabu neighbor as $new\_solution$\;
 }
 Continue the TS from this $new\_solution$\;
 Update the TS list by setting the selected entry operation/machine with the sum of the threshold $T_u$ plus the current iteration, increasing the number of iterations\;
 }
 Return the best $new\_solution$ as new $smart\_cell$\;
\caption{General description of the TS}
\label{alg:TS}
\end{algorithm}

\begin{figure}[htbp]
\begin{center}
\includegraphics[scale=0.8]{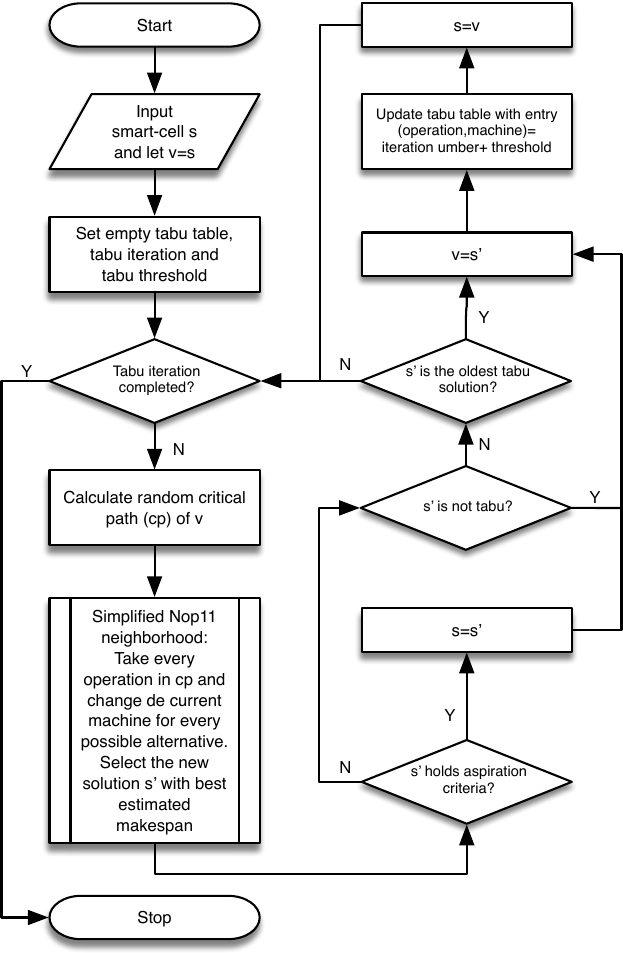}
\caption{Tabu search applied to the local search process of the GLNSA.}
\label{fig:DF_TS}
\end{center}
\end{figure}

The flow chart of the TS is shown in Fig. \ref{fig:DF_TS}.

\subsection{Critical path}

The tabu search for this job uses the following definition of the critical path $cp$. To form $cp$, one of the operations with completion time equal to the makespan is selected randomly. Once this operation has been chosen, the previous operation that precedes it is selected either on the same machine or by the previous operation of the same job, and the one whose completion time is equal to the initial time of the current operation is selected. If both previous operations have the same completion time, one of them is selected randomly. This process is repeated until an operation with start time $0$ is selected. These operations form a critical path $cp$ of length $q$.


\subsection{Simplified Nopt1 neighborhood}

The neighborhood used for the local search is based on the one defined in \cite{mastrolilli2000effective} as $Nopt1$. Given a critical path $cp$, in the original neighborhood $Nopt1$, for each operation in $cp$ a set of preceding and succeeding operations can be found in each feasible machine, such that a new placement of the operation between these operations on the new machine optimizes the makespan. The calculation of $Nopt1$ depends on the review of the start and tail times of the operations programmed in each machine, implying an almost constant computational time. When this operation is performed several times (such as in a meta-heuristic algorithm), the computational time can increase considerably, especially if the number of jobs and machines is large, and the system has high flexibility.


In this work, it is proposed to use a simplification of the Nopt1 neighborhood, where a search for the best position is not carried out to accommodate each operation in a new feasible machine, but its machine assignment is just changed in the string $ MS $, preserving the position in the string $ OS $. The idea is that the global search operations applied in the neighborhood based on cellular automata (insertion, swapping, and path relinking) are capable of finding this optimal assignment as the optimization algorithm advances, especially for systems with greater flexibility, where more machines are capable of processing the same operation. Thus, the objective is to avoid carrying out the operations that explicitly seek to accommodate an operation on a machine and directly take the position that is being refined by the global neighborhood's operations.

There are more optimal job placement positions for systems with high flexibility, given the greater availability of feasible machines, so this simplification focuses on showing that good results can be obtained with a reduced process for this type of FJSP instances.


\subsection{Parameters of the GLNSA}
 
The following are the parameters of the proposed algorithm:
 
\begin{itemize}
\item Number of iterations for the whole optimization process: $G_n$.
\item Number of $smart\_cells$: $S_n$.
\item Global neighborhood size: $l$.
\item Probabilities $\alpha_I$, $\alpha_S$ and $\alpha_P$ for insertion, swapping and path relinking, and probability $\alpha_M$ for mutation. 
\item Maximum number of stagnation iterations: $S_b$. 
\item Proportion of elitist solutions: $E_p$. 
\item Number of tabu iterations for every optimization iteration: $T_n$. 
\item Tabu threshold: $T_u$.
 \end{itemize}
 
 \subsection{Parameter tuning}


A preliminary study was carried out considering different values of the GLNSA parameters. They were applied for the same problem using a similar number of iterations to select the best parameters applied to instances of FJSP with high flexibility.

For the population size $S_n$, the number of iterations $G_n$, and the mutation probability $\alpha_M $, the results presented in \cite{gonzalez2015scatter} and \cite{li2016effective} are taken as a basis, since they are recent works that show great effectiveness both in the makespan calculation as well as in the runtime for FJSP instances.

For $G_n $, values of $200$ and $250$ were tested. Also, while $S_n$ is taken between $20$ and $100$ in \cite{gonzalez2015scatter} and defined as $400$ in \cite{li2016effective}, we tested $S_n$ between $40$ and $80$.

The number of neighbors $l$ that each $smart\_cell $ has in our algorithm to generate the global search neighborhood was tested with values between $2$ and $3$, in order to preserve a population close to $250$ solutions at most (number of $smart\_cells$ by number of neighbors) and keep a computational execution close to the cited references. To form the global neighborhood, the probability combinations $ (\alpha_I, \alpha_S, \alpha_P) $ with values $ (0.5,0.25,0.25) $, $ (0.25,0.5,0.25) $ and $ (0.25,0.25, 0.5)$ were tested. The mutation probability $ \alpha_M $ was taken with values $ 0.1 $ and $ 0.2 $.

To control the stagnation limit $S_b$, the value proposed by \cite{li2016effective} is taken to test $S_b$ with values $20$ and $40$, and for the elitist proportion of solutions, a value of $E_p$ of $0.025$ and $0.05$ is considered.

Without a doubt, the tabu search is the most computationally expensive process that the proposed algorithm has. In \cite{gonzalez2015scatter}, the execution of the TS is tested up to $10,000$ iterations per solution. In \cite{li2016effective}, this number goes up to about $80,000$ times per solution, of course, with different ways of creating solutions and estimating the makespan. Our algorithm uses the makespan estimation developed in \cite{mastrolilli2000effective} during the tabu search to reduce the computational time of the optimization process.

In our algorithm, we take a point of view similar to \cite{li2016effective}, using an increasing number of iterations of TS, as the number of iterations of the optimization process grows. For each iteration $i$, $T_n*i$ TS iterations will be taken for each $smart\_cell $, where $ T_n $ values of $ 1 $ and $ 2 $ are tested to keep a maximum TS close to $ 60,000 $ iterations per $smart\_cell $. Altogether, it took $384$ different combinations of parameters to tune the GLNSA.

For this preliminary tuning study, we took the problem $la31$ with more flexibility (known as part of the vdata set) initially proposed by \cite{hurink1994tabu} and downloaded from http://people.idsia.ch/\~{}monaldo/fjsp.html. From the tuning study, the parameters $ G_n = 250 $, $ S_n = 40 $, $ l = 2 $, $ \alpha_I = 0.5 $, $ \alpha_S = 0.25 $, $ \alpha_P = 0.25 $, $ \alpha_M = 0.1 $, $S_b = 40 $; $E_p = 0.025 $ and $ T_n = 1 $ were selected as the most appropriate values to apply the GLNSA.


The threshold used in the tabu list for each entry (operation/machine) consists of the sum of the length of the random critical path plus the number of feasible machines that can perform the corresponding operation. This criterion was also proposed in \cite{mastrolilli2000effective} and has been widely used by similar algorithms.

\begin{figure}[htbp]
\begin{center}
\includegraphics[scale=0.55]{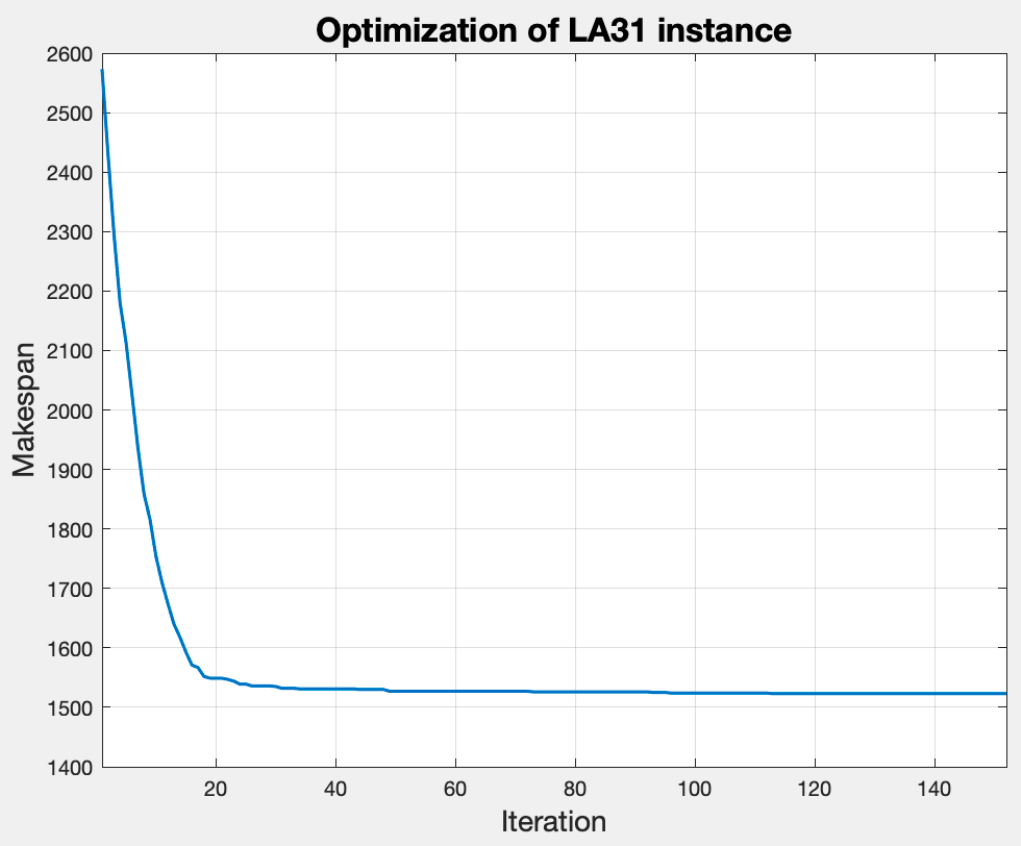}
\caption{Behavior of the makespan with tuned parameters for the $la31$-vdata instance.}
\label{fig:Ejemplo_Convergencia_la31}
\end{center}
\end{figure}

Figure \ref{fig:Ejemplo_Convergencia_la31} shows the convergence of the makespan by applying the GLNSA with the parameters indicated above on the instance $la31$-vdata. The implementation was developed in Matlab (the implementation characteristics and computational experiments are specified in detail in the next section), and the execution time was $56$ seconds.

\section{Experimental results}
\label{secc:Experiment}

The GLNSA was implemented in Matlab R2015a (TM) on a 2.3 GHz Intel Xeon W machine and 128 GB of RAM. Two sets of experiments from the HU \cite{hurink1994tabu} benchmark were taken to illustrate the effectiveness of the GLNSA. The first set takes $43$ instances from the HU-rdata benchmark, whose rate $\beta$=(flexibility/number of machines) is $\leq 0.4$. The second experiment takes $ 43 $ instances of the HU-vdata benchmark with the rate $\ beta = 0.5$, thus bringing a total of $ 86 $ instances of the FJSP. The rate $\beta$ is between $ 0 $ and $ 1 $. A higher value indicates that more machines can perform more different operations. This way, a value $ \beta = 0.5$ indicates that an operation can be processed by around half the machines, already implying a high degree of flexibility. Since the objective is to minimize the makespan, the results of GLNSA are compared with other state-of-art algorithms, comparing with the solutions reported in \cite{li2016effective}, since the algorithm of this reference obtains outstanding makespan values and the lowest computational time reported in our knowledge.

\subsection{First experiment, HU-rdata instances}

Table \ref{tabla:HU-rdata} shows the comparison of GLNSA with other algorithms, $n$ and $m$ indicate the number of jobs and the number of machines respectively, and the third column shows the flexibility rate $\beta$. The algorithms TSN1 and TSN2 are the methods proposed in \cite{hurink1994tabu}, IATS is the method reported by \cite{dauzere1997integrated}, TS is the algorithm developed in \cite{mastrolilli2000effective}, and HA is the process described by \cite{li2016effective}, which also reports the best execution times. In this sense, the time of the HA and GLNSA algorithms is reported in seconds (s). The results marked with $*$ are the best obtained among the $6$ algorithms.

\begin{table}[thbp]
\centering
\tiny
\caption{\label{tabla:HU-rdata}Experimental results for the HU-rdata instances.} 
\begin{tabular}{lllllllllll}
\hline
Instance & n $\times$ m & $\beta$ & TSN1 & TSN2 & IATS & TS & HA & HA(s) & GLNSA & GLNSA(s) \\ 
\hline
mt06 & 6 $\times$ 6 & $0.33$  & $47*$ & $47*$ & $47*$ & $47*$ & $47*$ & $0.01$ & $47*$ & $0.49$  \\
mt10 & 10 $\times$ 10 & $0.2$ &  $737$ &  $737$ &  $686*$ &  $686*$ &  $686*$ &  $2.42$ &  $686*$ &  $3.77$ \\
mt20 & 20 $\times$ 5 & $0.4$ &  $1028$ &  $1028$ &  $1024$ &  $1022*$ &  $1022*$ &  $12.79$ &  $1022*$ &  $10.45$  \\
la01 & 10 $\times$ 5 & $0.4$ &  $577$ &  $577$ &  $574$ &  $571$ &  $570*$ &  $13.31$ &  $571$ &  $10.66$  \\
la02 & 10 $\times$ 5 & $0.4$ &  $535$ &  $535$ &  $532$ &  $530*$ &  $530*$ &  $3.84$ &  $530*$ &  $2.11$  \\
la03 & 10 $\times$ 5 & $0.4$ &  $481$ &  $486$ &  $479$ &  $478$ &  $477*$ &  $4.85$ &  $477*$ &  $2.82$  \\
la04 & 10 $\times$ 5 & $0.4$ &  $509$ &  $506$ &  $504$ &  $502*$ &  $502*$ &  $4.13$ &  $502*$ &  $3.33$  \\
la05 & 10 $\times$ 5 & $0.4$ &  $460$ &  $458$ &  $458$ &  $457*$ &  $457*$ &  $4.23$ &  $457*$ &  $2.72$  \\
la06 & 15 $\times$ 5 & $0.4$ &  $801$ &  $803$ &  $800$ &  $799*$ &  $799*$ &  $3.54$ &  $799*$ &  $3.09$  \\
la07 & 15 $\times$ 5 & $0.4$ &  $752$ &  $752$ &  $750$ &  $750$ &  $749*$ &  $7.35$ &  $749*$ &  $4.11$  \\
la08 & 15 $\times$ 5 & $0.4$ &  $767$ &  $768$ &  $767$ &  $765*$ &  $765*$ &  $9.89$ &  $765*$ &  $4.16$  \\
la09 & 15 $\times$ 5 & $0.4$ &  $859$ &  $857$ &  $854$ &  $853*$ &  $853*$ &  $11.93$ &  $853*$ &  $4.02$  \\
la10 & 15 $\times$ 5 & $0.4$ &  $806$ &  $805$ &  $805$ &  $804*$ &  $804*$ &  $8.88$ &  $804*$ &  $3.79$  \\
la11 & 20 $\times$ 5 & $0.4$ &  $1073$ &  $1073$ &  $1072$ &  $1071*$ &  $1071*$ &  $2.73$ &  $1071*$ &  $2.41$  \\
la12 & 20 $\times$ 5 & $0.4$ &  $937$ &  $937$ &  $936*$ &  $936*$ &  $936*$ &  $2.12$ &  $936*$ &  $2.34$  \\
la13 & 20 $\times$ 5 & $0.4$ &  $1039$ &  $1039$ &  $1038*$ &  $1038*$ &  $1038*$ &  $3.15$ &  $1038*$ &  $2.36$  \\
la14 & 20 $\times$ 5 & $0.4$ &  $1071$ &  $1071$ &  $1070*$ &  $1070*$ &  $1070*$ &  $2.67$ &  $1070*$ &  $2.66$  \\
la15 & 20 $\times$ 5 & $0.4$ &  $1093$ &  $1093$ &  $1090$ &  $1090$ &  $1090$ &  $9.73$ &  $1089*$ &  $3.43$  \\
la16 & 10 $\times$ 10 & $0.2$ &  $717*$ &  $717*$ &  $717*$ &  $717*$ &  $717*$ &  $0.98$ &  $717*$ &  $1.46$  \\
la17 & 10 $\times$ 10 & $0.2$ &  $646*$ &  $646*$ &  $646*$ &  $646*$ &  $646*$ &  $0.58$ &  $646*$ &  $1.88$  \\
la18 & 10 $\times$ 10 & $0.2$ &  $674$ &  $673$ &  $669$ &  $666*$ &  $666*$ &  $1.12$ &  $666*$ &  $3.16$  \\
la19 & 10 $\times$ 10 & $0.2$ &  $725$ &  $709$ &  $703$ &  $700*$ &  $700*$ &  $1.6*$ &  $700*$ &  $3.07$  \\
la20 & 10 $\times$ 10 & $0.2$ &  $756*$ &  $756*$ &  $756*$ &  $756*$ &  $756*$ &  $0.59$ &  $756*$ &  $1.14$  \\
la21 & 15 $\times$ 10 & $0.2$ &  $861$ &  $861$ &  $846$ &  $835*$ &  $835*$ &  $13.88$ &  $852$ &  $13.32$  \\
la22 & 15 $\times$ 10 & $0.2$ &  $790$ &  $795$ &  $772$ &  $760*$ &  $760*$ &  $10.31$ &  $774$ &  $13.84$  \\
la23 & 15 $\times$ 10 & $0.2$ &  $884$ &  $887$ &  $853$ &  $842$ &  $840*$ &  $25.04$ &  $854$ &  $16.52$  \\
la24 & 15 $\times$ 10 & $0.2$ &  $825$ &  $830$ &  $820$ &  $808$ &  $806*$ &  $22.32$ &  $826$ &  $11.21$  \\
la25 & 15 $\times$ 10 & $0.2$ &  $823$ &  $821$ &  $802$ &  $791$ &  $789*$ &  $25.09$ &  $803$ &  $19.79$  \\
la26 & 20 $\times$ 10 & $0.2$ &  $1086$ &  $1087$ &  $1070$ &  $1061*$ &  $1061*$ &  $38.35$ &  $1075$ &  $28.14$  \\
la27 & 20 $\times$ 10 & $0.2$ &  $1109$ &  $1115$ &  $1100$ &  $1091$ &  $1089*$ &  $56.81$ &  $1109$ &  $32.49$  \\
la28 & 20 $\times$ 10 & $0.2$ &  $1097$ &  $1090$ &  $1085$ &  $1080$ &  $1079*$ &  $55.02$ &  $1096$ &  $37.52$  \\
la29 & 20 $\times$ 10 & $0.2$ &  $1016$ &  $1017$ &  $1004$ &  $998$ &  $997*$ &  $52.49$ &  $1008$ &  $44.37$  \\
la30 & 20 $\times$ 10 & $0.2$ &  $1105$ &  $1108$ &  $1089$ &  $1078*$ &  $1078*$ &  $52.88$ &  $1096$ &  $45.87$  \\
la31 & 30 $\times$ 10 & $0.2$ &  $1532$ &  $1533$ &  $1528$ &  $1521*$ &  $1521*$ &  $50.89$ &  $1527$ &  $39.76$  \\
la32 & 30 $\times$ 10 & $0.2$ &  $1668$ &  $1668$ &  $1660$ &  $1659*$ &  $1659*$ &  $43.56$ &  $1667$ &  $42.66$  \\
la33 & 30 $\times$ 10 & $0.2$ &  $1511$ &  $1507$ &  $1501$ &  $1499*$ &  $1499*$ &  $50.90$ &  $1504$ &  $42.72$  \\
la34 &30 $\times$ 10 & $0.2$ &  $1542$ &  $1543$ &  $1539$ &  $1536$ &  $1536*$ &  $38.98$ &  $1540$ &  $49.27$  \\
la35 & 30 $\times$ 10 & $0.2$ &  $1559$ &  $1559$ &  $1555$ &  $1550*$ &  $1550*$ &  $45.19$ &  $1555$ &  $47.05$  \\
la36 & 15 $\times$ 15 & $0.13$ &  $1054$ &  $1071$ &  $1030$ &  $1030$ &  $1028*$ &  $56.64$ &  $1053$ &  $38.40$  \\
la37 & 15 $\times$ 15 & $0.13$ &  $1122$ &  $1132$ &  $1082$ &  $1077$ &  $1074*$ &  $55.47$ &  $1093$ &  $43.17$  \\
la38 & 15 $\times$ 15 & $0.13$ &  $1004$ &  $1001$ &  $989$ &  $962$ &  $960*$ &  $43.71$ &  $999$ &  $32.08$  \\
la39 & 15 $\times$ 15 & $0.13$ &  $1041$ &  $1068$ &  $1024*$ &  $1024*$ &  $1024*$ &  $9.28$ &  $1034$ &  $15.57$  \\
la40 & 15 $\times$ 15 & $0.13$ &  $1009$ &  $1009$ &  $980$ &  $970*$ &  $970*$ &  $43.40$ &  $997$ &  $30.69$  \\
\hline
\end{tabular}
\end{table}

In Table \ref{tabla:HU-rdata}, we can see that the proposed GLNSA obtains the best results in almost all the problems with the highest value for $\beta$, and achieves the best results in $5$ of the problems with less value for $\beta$, as expected, given the low flexibility of the machines represented in these instances ($16$ to $40$).

It should be noted that, for instance $la15$, the GLNSA achieves the best makespan, with a value of $ 1089 $. Compared to the HA, the GLNSA has a lower execution time at $33$ of the $43$ instances in this experiment. Based on the results of Table \ref{tabla:HU-rdata}, it is shown that GLNSA has an efficiency comparable to HA for instances with the highest values of $\beta$ and a lower computational time in most of these cases. Below are the Gantt charts of some of the job schedules obtained by the GLNSA.

\begin{figure}[htbp]
\begin{center}
\includegraphics[scale=0.33]{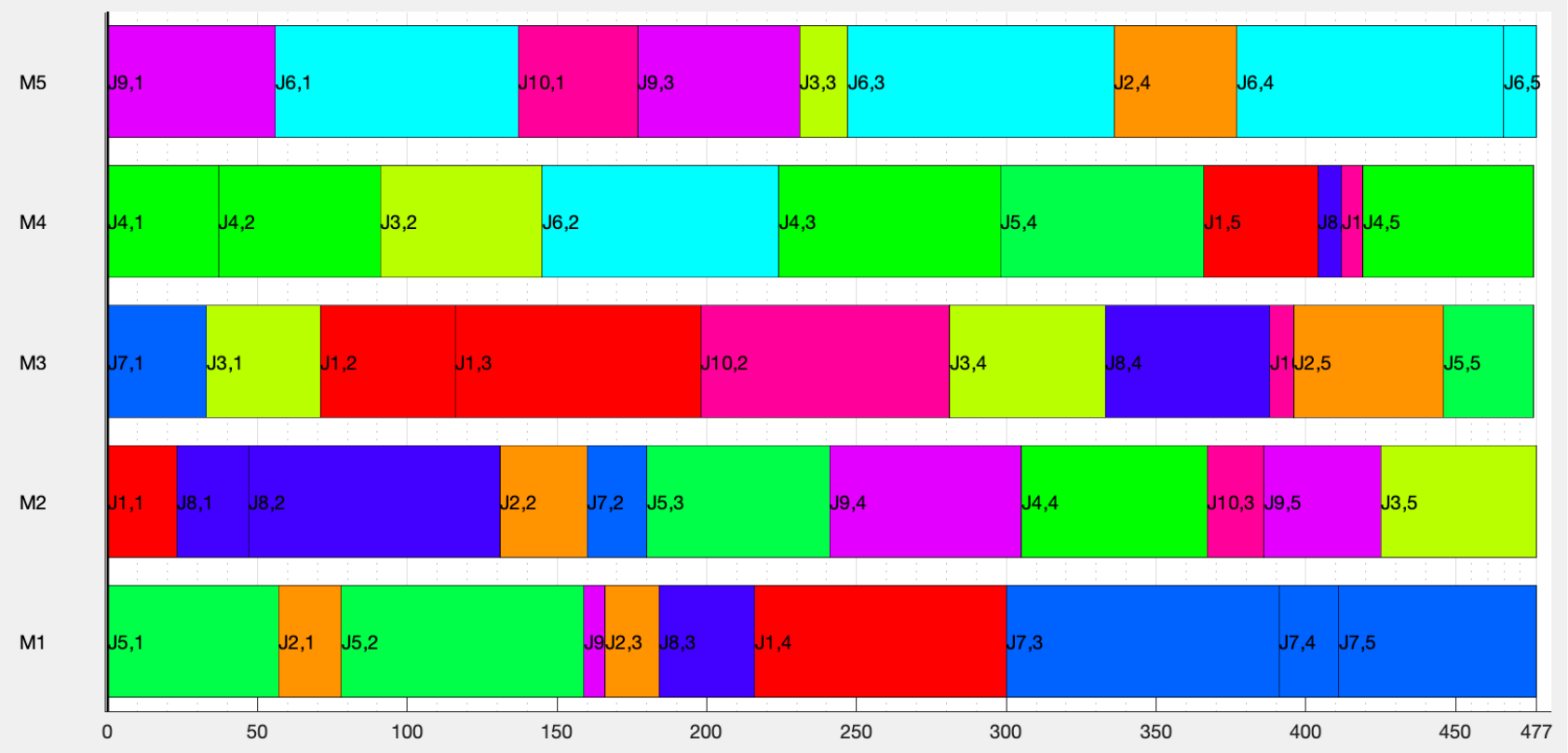}
\caption{Gantt chart of the solution obtained for the instance $la03$-rdata with a makespan of $477$.}
\label{fig:rdata-la03-477}
\end{center}
\end{figure}

\begin{figure}[htbp]
\begin{center}
\includegraphics[scale=0.26]{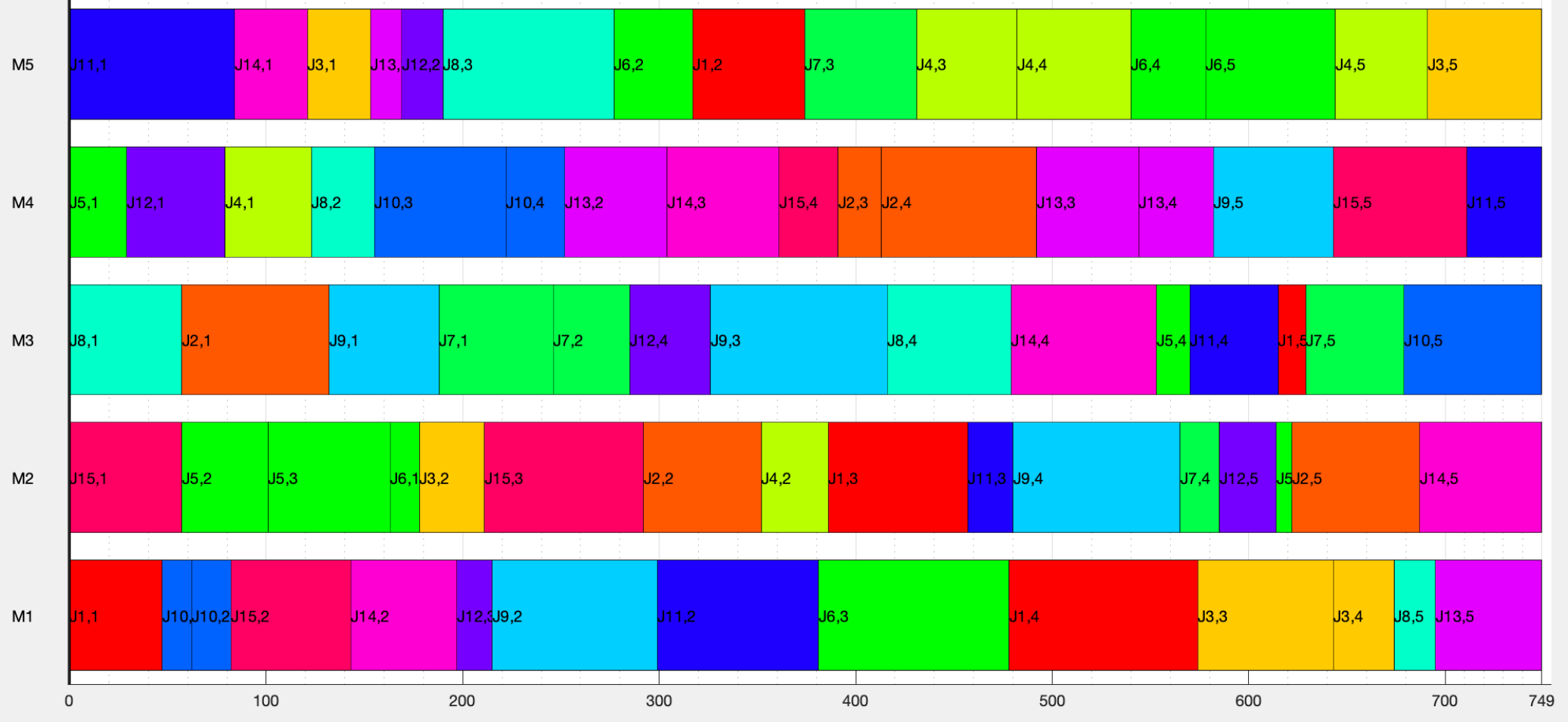}
\caption{Gantt chart of the solution obtained for the instance $la07$-rdata with a makespan of $749$.}
\label{fig:rdata-la07-749}
\end{center}
\end{figure}

\begin{figure}[htbp]
\begin{center}
\includegraphics[scale=0.4]{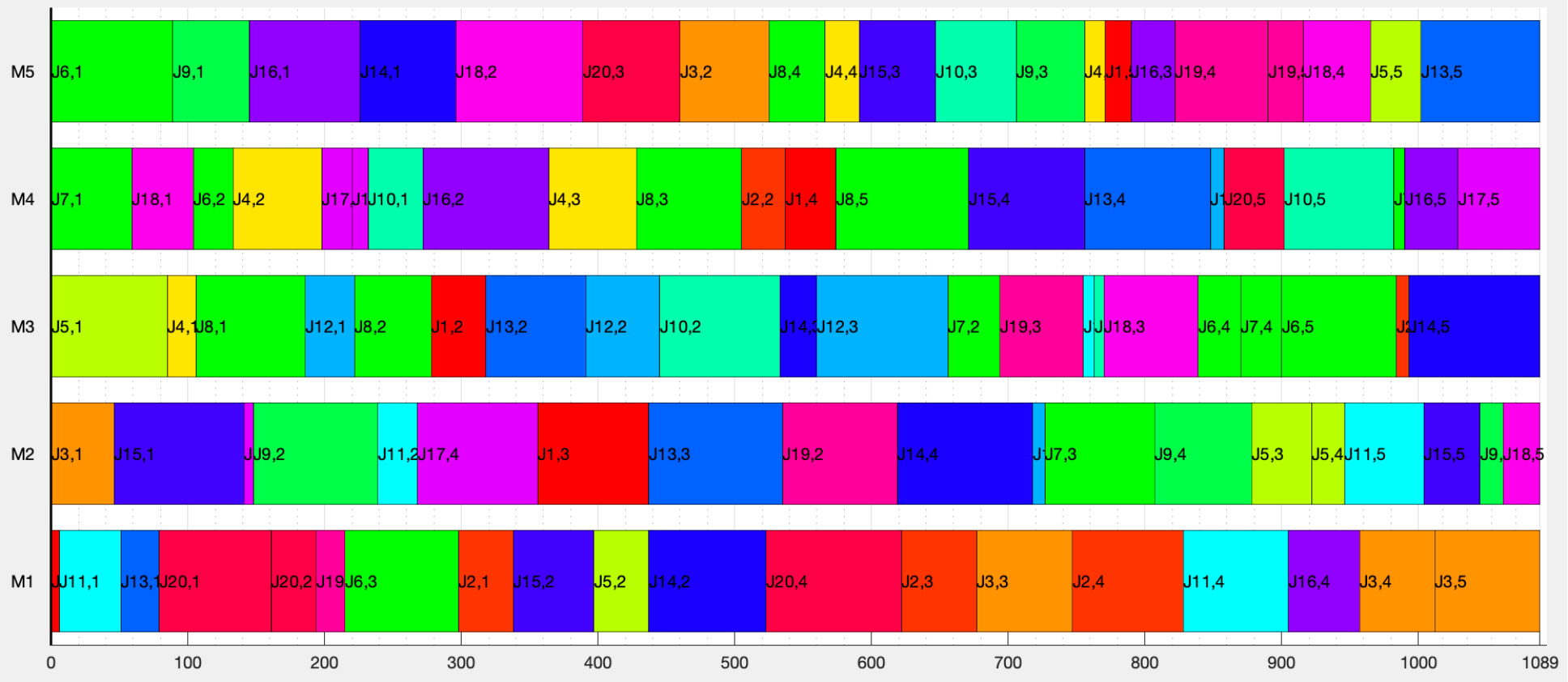}
\caption{Gantt chart of the solution obtained for the instance $la15$-rdata with a makespan of $1089$.}
\label{fig:rdata-la15-1089}
\end{center}
\end{figure}

\begin{figure}[htbp]
\begin{center}
\includegraphics[scale=0.415]{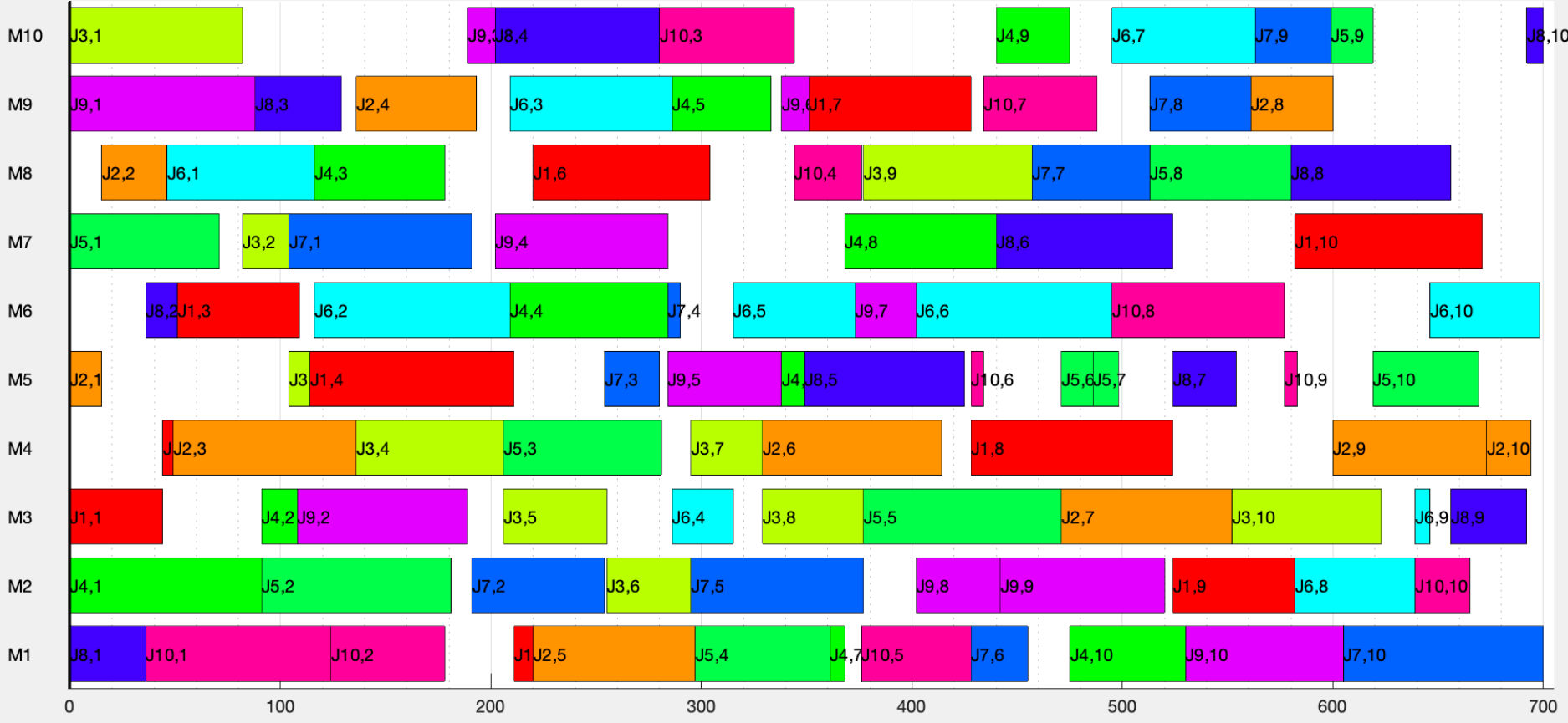}
\caption{Gantt chart of the solution obtained for the instance $la19$-rdata with a makespan of $700$.}
\label{fig:rdata-la19-700}
\end{center}
\end{figure}

\begin{figure}[htbp]
\begin{center}
\includegraphics[scale=0.33]{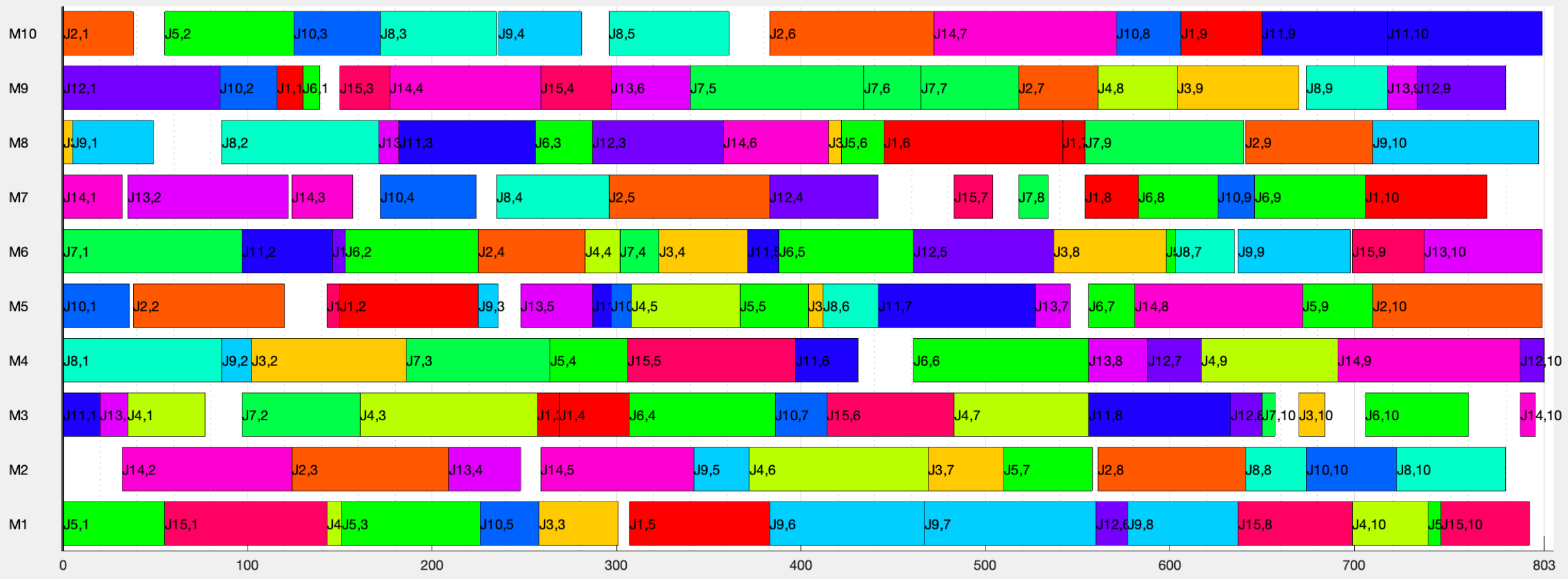}
\caption{Gantt chart of the solution obtained for the instance $la25$-rdata with a makespan of $803$.}
\label{fig:rdata-la25-803}
\end{center}
\end{figure}

\begin{figure}[htbp]
\begin{center}
\includegraphics[scale=0.38]{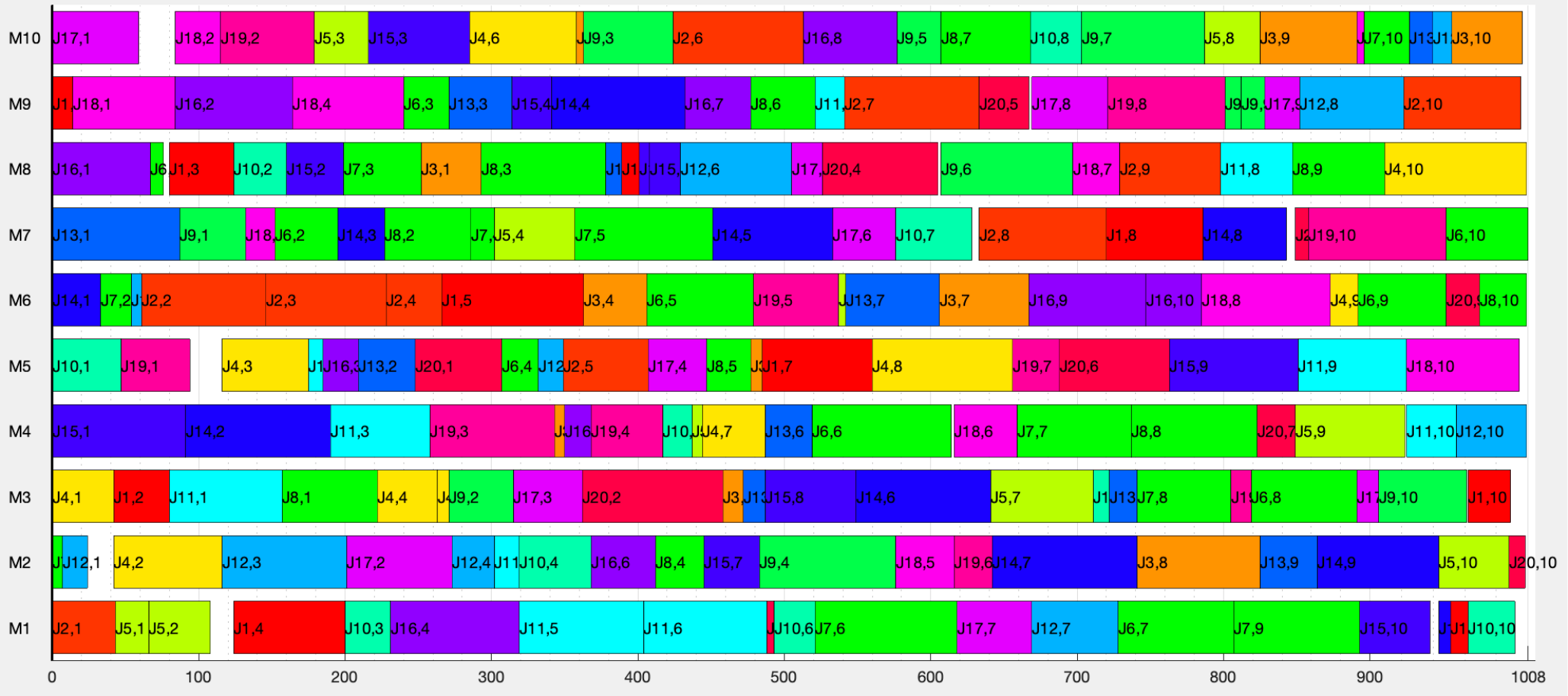}
\caption{Gantt chart of the solution obtained for the instance $la29$-rdata with a makespan of $1008$.}
\label{fig:rdata-la29-1008}
\end{center}
\end{figure}

\begin{figure}[htbp]
\begin{center}
\includegraphics[scale=0.29]{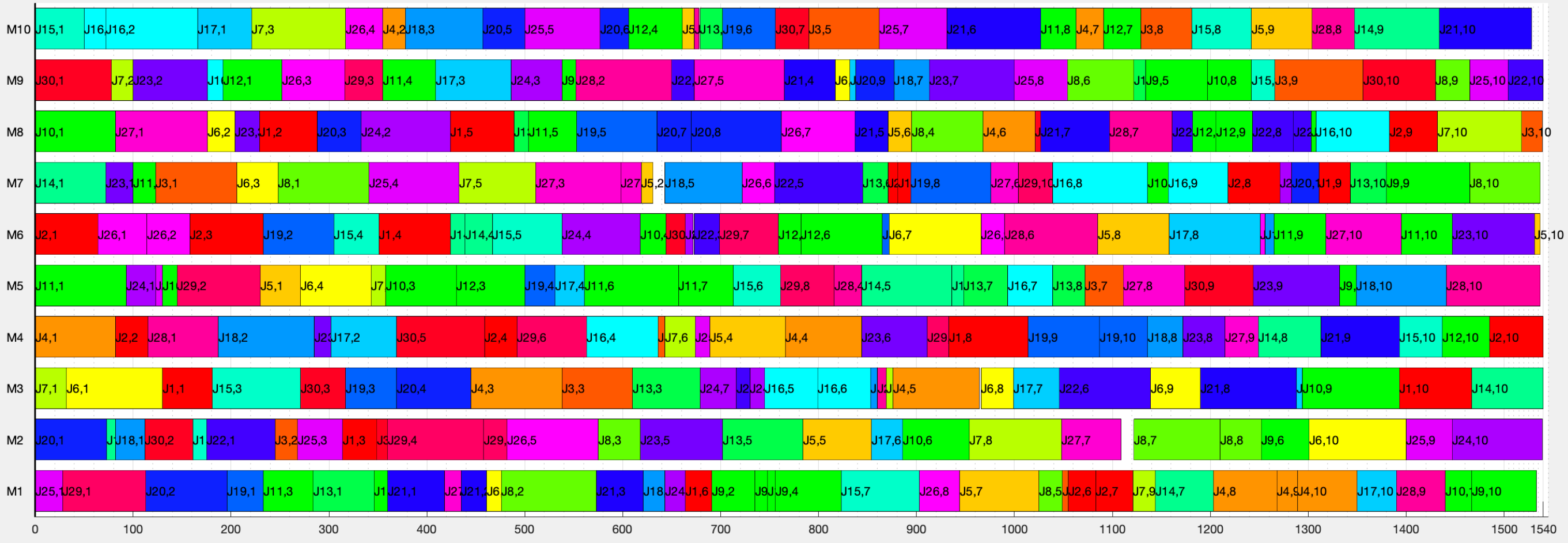}
\caption{Gantt chart of the solution obtained for the instance $la34$-rdata with a makespan of $1540$.}
\label{fig:rdata-la34-1540}
\end{center}
\end{figure}

\begin{figure}[htbp]
\begin{center}
\includegraphics[scale=0.33]{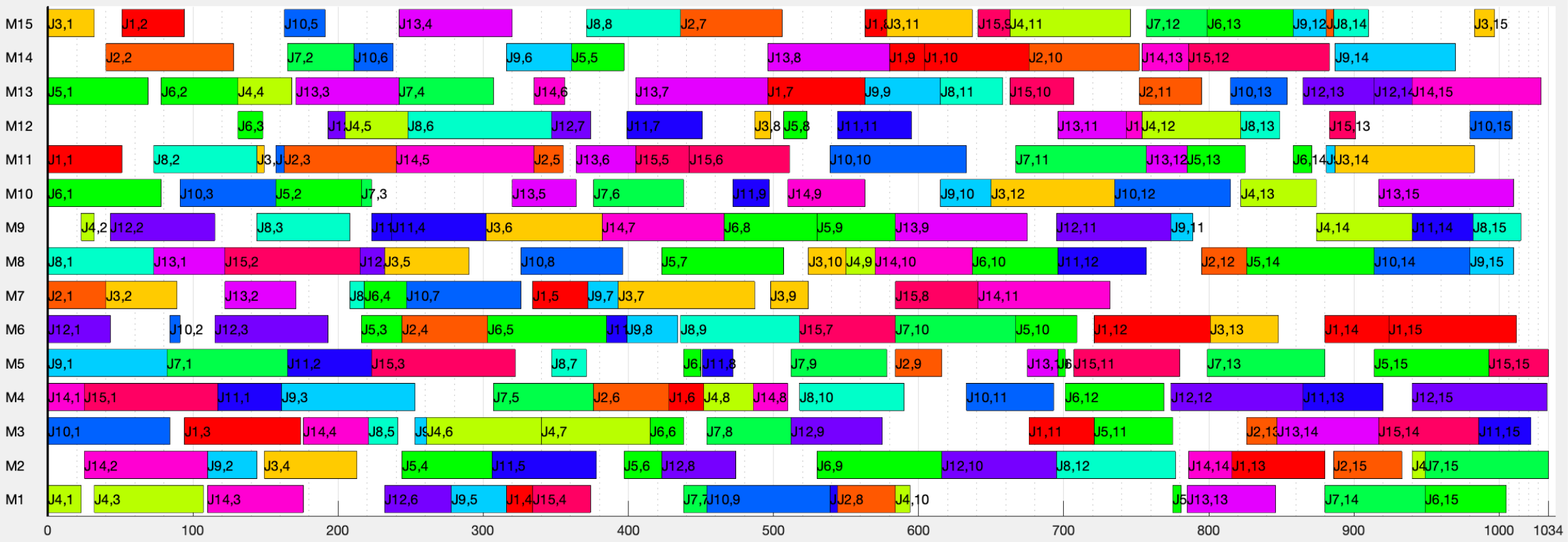}
\caption{Gantt chart of the solution obtained for the instance $la39$-rdata with a makespan of $1034$.}
\label{fig:rdata-la39-1034}
\end{center}
\end{figure}


\subsection{Second experiment, HU-vdata instances}

\begin{table}[th]
\centering
\tiny
\caption{\label{tabla:HU-vdata}Experimental results for the HU-vdata instances.} 
\begin{tabular}{lllllllllll}
\hline
Problema & n $\times$ m & $\beta$ & TSN1 & TSN2 & IATS & TS & HA & HA(s) & GLNSA & GLNSA(s) \\ 
\hline
mt06 & 6 $\times$ 6 & $0.5$  & $47*$ & $47*$ & $47*$ & $47*$ & $47*$ & $0.01$ & $47*$ & $0.44$  \\
mt10 & 10 $\times$ 10 & $0.5$ & $655*$ & $655*$ & $655*$ & $655*$ & $655*$ & $0.22$ & $655*$ & $0.91$  \\
mt20 & 20 $\times$ 5 & $0.5$ & $1023$ & $1023$ & $1022*$ & $1022*$ & $1022*$ & $5.91$ & $1022*$ & $4.02$  \\
la01 & 10 $\times$ 5 & $0.5$ & $573$ & $575$ & $572$ & $570*$ & $570*$ & $0.92$ & $570*$ & $0.81$  \\
la02 & 10 $\times$ 5 & $0.5$ & $531$ & $530$ & $529*$ & $529*$ & $529*$ & $1.23$ & $529*$ & $1.19$  \\
la03 & 10 $\times$ 5 & $0.5$ & $482$ & $481$ & $479$ & $477*$ & $477*$ & $1.29$ & $477*$ & $1.16$  \\
la04 & 10 $\times$ 5 & $0.5$ & $504$ & $503$ & $503$ & $502*$ & $502*$ & $1.09$ & $502*$ & $1.09$  \\
la05 & 10 $\times$ 5 & $0.5$ & $464$ & $461$ & $460$ & $457*$ & $457*$ & $10.47$ & $457*$ & $1.88$  \\
la06 & 15 $\times$ 5 & $0.5$ & $802$ & $799*$ & $800$ & $799*$ & $799*$ & $2.95$ & $799*$ & $1.96$  \\
la07 & 15 $\times$ 5 & $0.5$ & $751$ & $752$ & $750$ & $749*$ & $749*$ & $14.38$ & $749*$ & $3.24$  \\
la08 & 15 $\times$ 5 & $0.5$ & $766$ & $766$ & $766$ & $765*$ & $765*$ & $16.75$ & $765*$ & $3.37$  \\
la09 & 15 $\times$ 5 & $0.5$ & $854$ & $854$ & $853*$ & $853*$ & $853*$ & $9.67$ & $853*$ & $3.18$  \\
la10 & 15 $\times$ 5 & $0.5$ & $805$ & $805$ & $805$ & $804*$ & $804*$ & $2.54$ & $804*$ & $1.88$  \\
la11 & 20 $\times$ 5 & $0.5$ & $1073$ & $1073$ & $1071*$ & $1071*$ & $1071*$ & $1.73$ & $1071*$ & $2.12$  \\
la12 & 20 $\times$ 5 & $0.5$ & $940$ & $940$ & $936*$ & $936*$ & $936*$ & $2.73$ & $936*$ & $2.08$  \\
la13 & 20 $\times$ 5 & $0.5$ & $1040$ & $1041$ & $1038*$ & $1038*$ & $1038*$ & $4.34$ & $1038*$ & $3.50$  \\
la14 & 20 $\times$ 5 & $0.5$ & $1071$ & $1080$ & $1070*$ & $1070*$ & $1070*$ & $3.38$ & $1070*$ & $3.17$  \\
la15 & 20 $\times$ 5 & $0.5$ & $1091$ & $1091$ & $1089*$ & $1089*$ & $1089*$ & $17.51$ & $1089*$ & $4.01$  \\
la16 & 10 $\times$ 10 & $0.5$ & $717*$ & $717*$ & $717*$ & $717*$ & $717*$ & $0.44$ & $717*$ & $0.79$  \\
la17 & 10 $\times$ 10 & $0.5$ & $646*$ & $646*$ & $646*$ & $646*$ & $646*$ & $0.56$ & $646*$ & $0.69$  \\
la18 & 10 $\times$ 10 & $0.5$ & $663*$ & $663*$ & $663*$ & $663*$ & $663*$ & $0.48$ & $663*$ & $0.93$  \\
la19 & 10 $\times$ 10 & $0.5$ & $617*$ & $617*$ & $617*$ & $617*$ & $617*$ & $1.22$ & $617*$ & $2.86$  \\
la20 & 10 $\times$ 10 & $0.5$ & $756*$ & $756*$ & $756*$ & $756*$ & $756*$ & $0.41$ & $756*$ & $0.73$  \\
la21 & 15 $\times$ 10 & $0.5$ & $826$ & $825$ & $814$ & $806$ & $804*$ & $44.19$ & $806$ & $37.83$  \\
la22 & 15 $\times$ 10 & $0.5$ & $745$ & $744$ & $744$ & $739$ & $738$ & $41.23$ & $737*$ & $33.04$  \\
la23 & 15 $\times$ 10 & $0.5$ & $826$ & $829$ & $818$ & $815$ & $813*$ & $35.37$ & $813*$ & $32.76$  \\
la24 & 15 $\times$ 10 & $0.5$ & $796$ & $796$ & $784$ & $777*$ & $777*$ & $38.89$ & $777*$ & $31.07$  \\
la25 & 15 $\times$ 10 & $0.5$ & $770$ & $769$ & $757$ & $756$ & $754*$ & $39.71$ & $754*$ & $33.02$  \\
la26 & 20 $\times$ 10 & $0.5$ & $1058$ & $1058$ & $1056$ & $1054$ & $1053*$ & $41.43$ & $1054$ & $33.36$  \\
la27 & 20 $\times$ 10 & $0.5$ & $1088$ & $1088$ & $1087$ & $1085*$ & $1085*$ & $35.29$ & $1085*$ & $25.44$  \\
la28 & 20 $\times$ 10 & $0.5$ & $1073$ & $1073$ & $1072$ & $1070*$ & $1070*$ & $42.17$ & $1070*$ & $33.66$  \\
la29 & 20 $\times$ 10 & $0.5$ & $995$ & $996$ & $997$ & $994*$ & $994*$ & $44.36$ & $994*$ & $34.60$  \\
la30 & 20 $\times$ 10 & $0.5$ & $1071$ & $1070$ & $1071$ & $1069*$ & $1069*$ & $51.31$ & $1069*$ & $47.37$  \\
la31 & 30 $\times$ 10 & $0.5$ & $1521$ & $1521$ & $1521$ & $1520*$ & $1520*$ & $58.25$ & $1520*$ & $55.86$  \\
la32 & 30 $\times$ 10 & $0.5$ & $1658*$ & $1659$ & $1659$ & $1658*$ & $1658*$ & $65.78$ & $1658*$ & $56.42$  \\
la33 & 30 $\times$ 10 & $0.5$ & $1498$ & $1499$ & $1498$ & $1497*$ & $1497*$ & $53.82$ & $1497*$ & $50.54$  \\
la34 &30 $\times$ 10 & $0.5$ & $1536$ & $1538$ & $1537$ & $1535*$ & $1535*$ & $62.55$ & $1535*$ & $51.05$  \\
la35 & 30 $\times$ 10 & $0.5$ & $1553$ & $1551$ & $1551$ & $1549*$ & $1549*$ & $67.02$ & $1549*$ & $50.26$  \\
la36 & 15 $\times$ 15 & $0.5$ & $948*$ & $948*$ & $948*$ & $948*$ & $948*$ & $4.57$ & $948*$ & $11.64$  \\
la37 & 15 $\times$ 15 & $0.5$ & $986*$ & $986*$ & $986*$ & $986*$ & $986*$ & $4.88$ & $986*$ & $17.45$  \\
la38 & 15 $\times$ 15 & $0.5$ & $943*$ & $943*$ & $943*$ & $943*$ & $943*$ & $2.13$ & $943*$ & $5.92$  \\
la39 & 15 $\times$ 15 & $0.5$ & $922*$ & $922*$ & $922*$ & $922*$ & $922*$ & $4.33$ & $922*$ & $17.36$  \\
la40 & 15 $\times$ 15 & $0.5$ & $955*$ & $955*$ & $955*$ & $955*$ & $955*$ & $4.61$ & $955*$ & $6.71$  \\
\hline
\end{tabular}
\end{table}

Table \ref{tabla:HU-vdata} shows that the proposed GLNSA obtains the best results for all problems except for $la21$. These results confirm the observation that the simplified Nopt1 neighborhood works adequately for problems with greater flexibility $\beta = 0.5$.

Again for this benchmark, the GLNSA achieves a better value than the rest for $la22$. Compared to the HA, the GLNSA has a lower execution time in $32$ of the $43$ instances. Based on the results of Table \ref{tabla:HU-rdata}, it is shown that GLNSA has an efficiency comparable to HA for instances with greater flexibility on their machines and with lesser computational time in most vdata instances.

\begin{figure}[htbp]
\begin{center}
\includegraphics[scale=0.38]{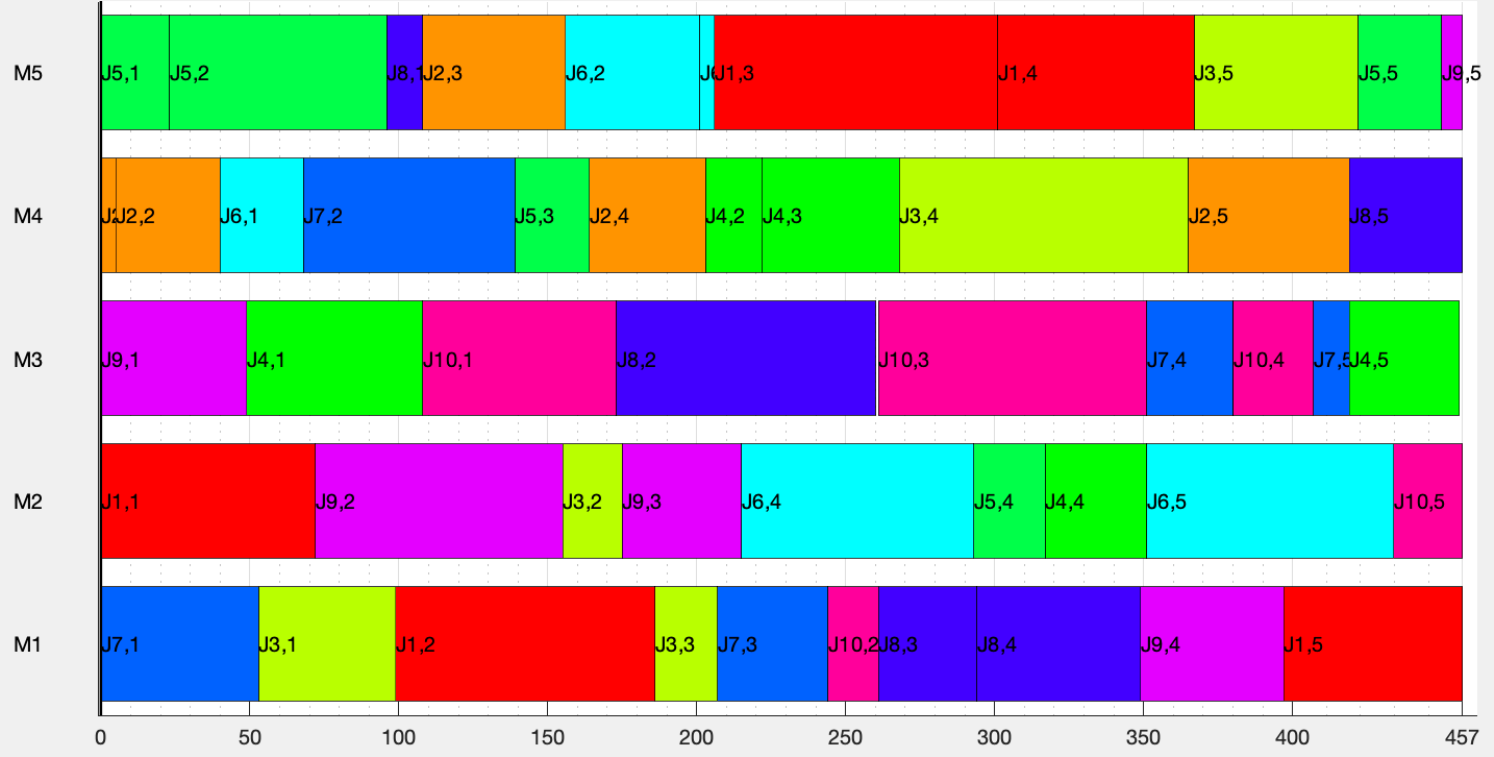}
\caption{Gantt chart of the solution obtained for the instance $la05$-vdata with a makespan of $457$.}
\label{fig:vdata-la05-457}
\end{center}
\end{figure}

\begin{figure}[htbp]
\begin{center}
\includegraphics[scale=0.37]{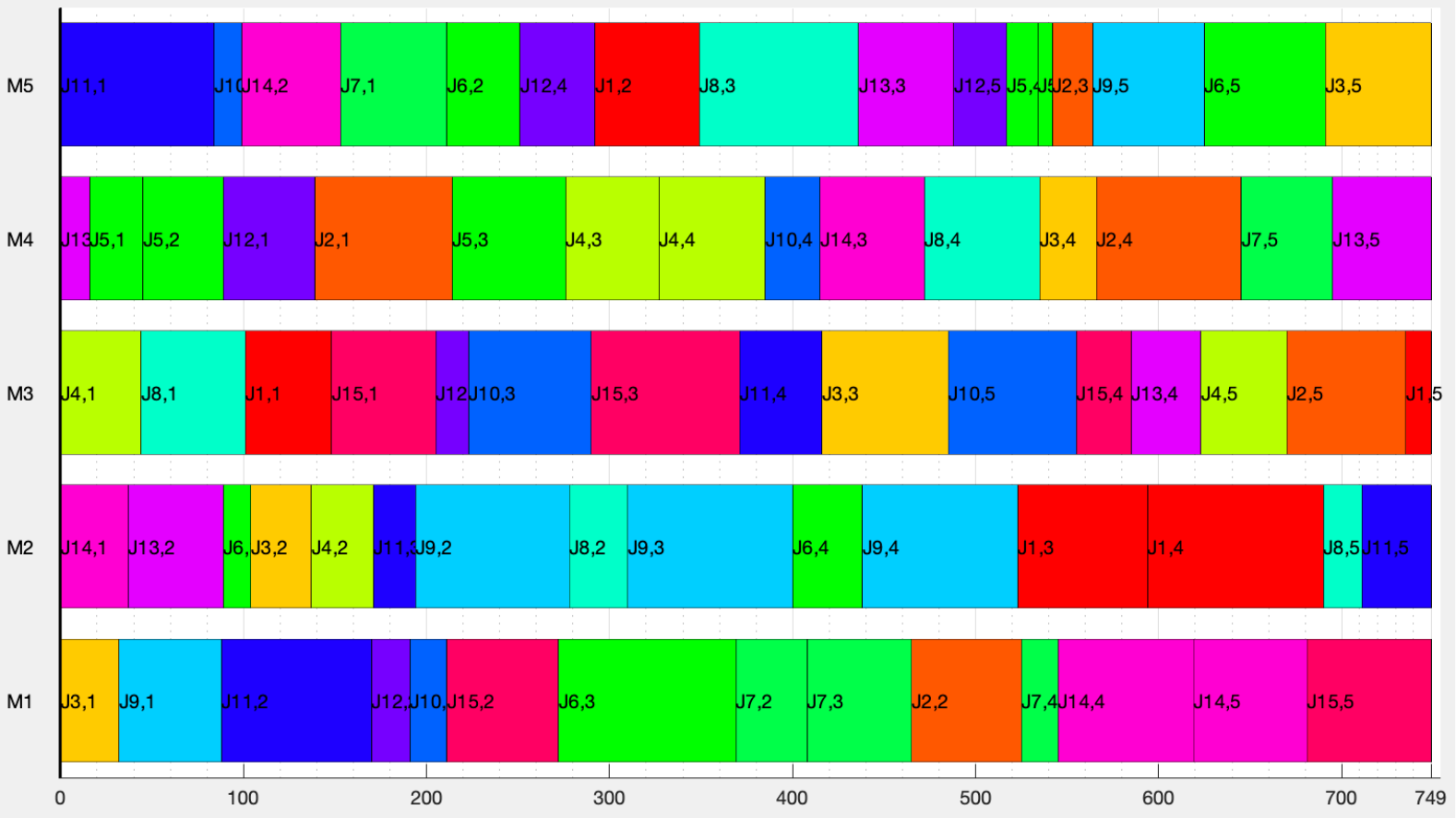}
\caption{Gantt chart of the solution obtained for the instance $la07$-vdata with a makespan of $749$.}
\label{fig:vdata-la07-749}
\end{center}
\end{figure}

\begin{figure}[htbp]
\begin{center}
\includegraphics[scale=0.45]{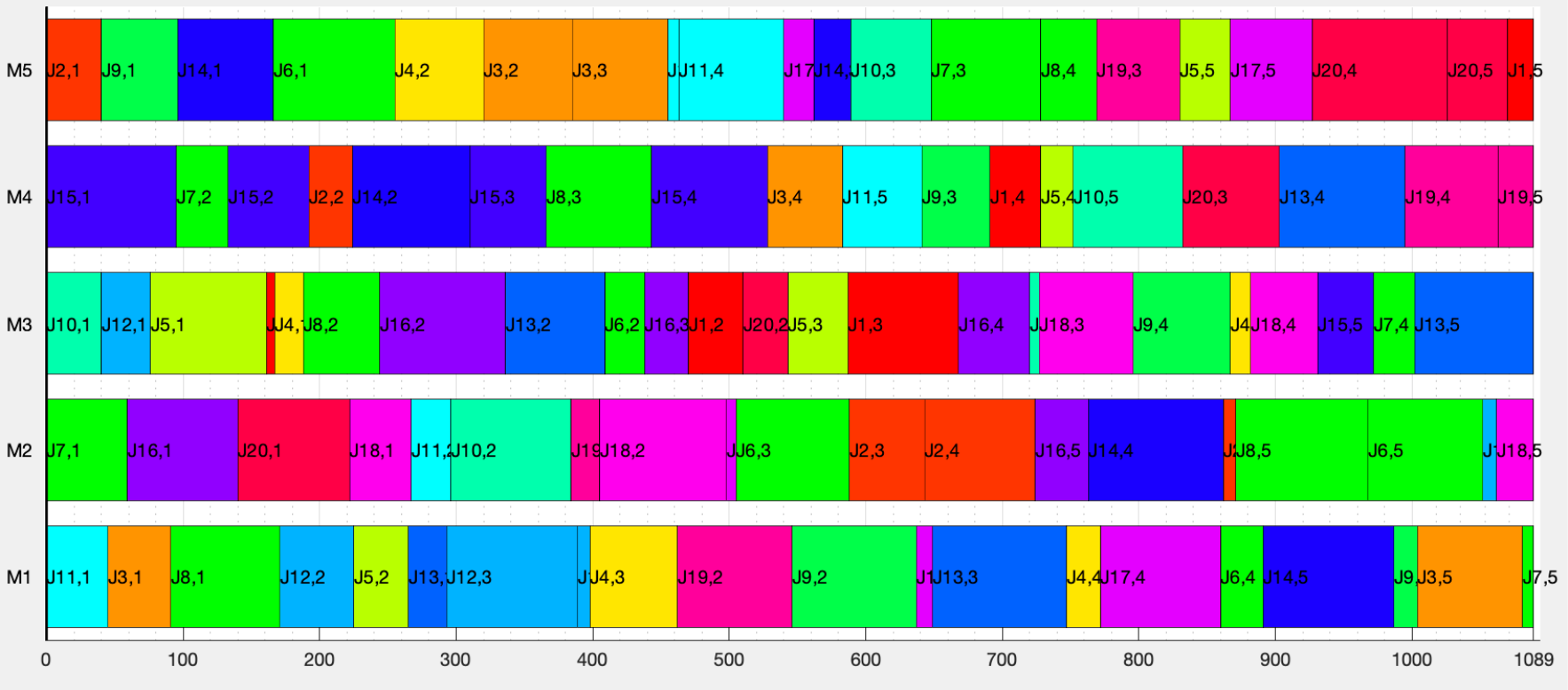}
\caption{Gantt chart of the solution obtained for the instance $la15$-vdata with a makespan of $1089$.}
\label{fig:vdata-la15-1089}
\end{center}
\end{figure}

\begin{figure}[htbp]
\begin{center}
\includegraphics[scale=0.45]{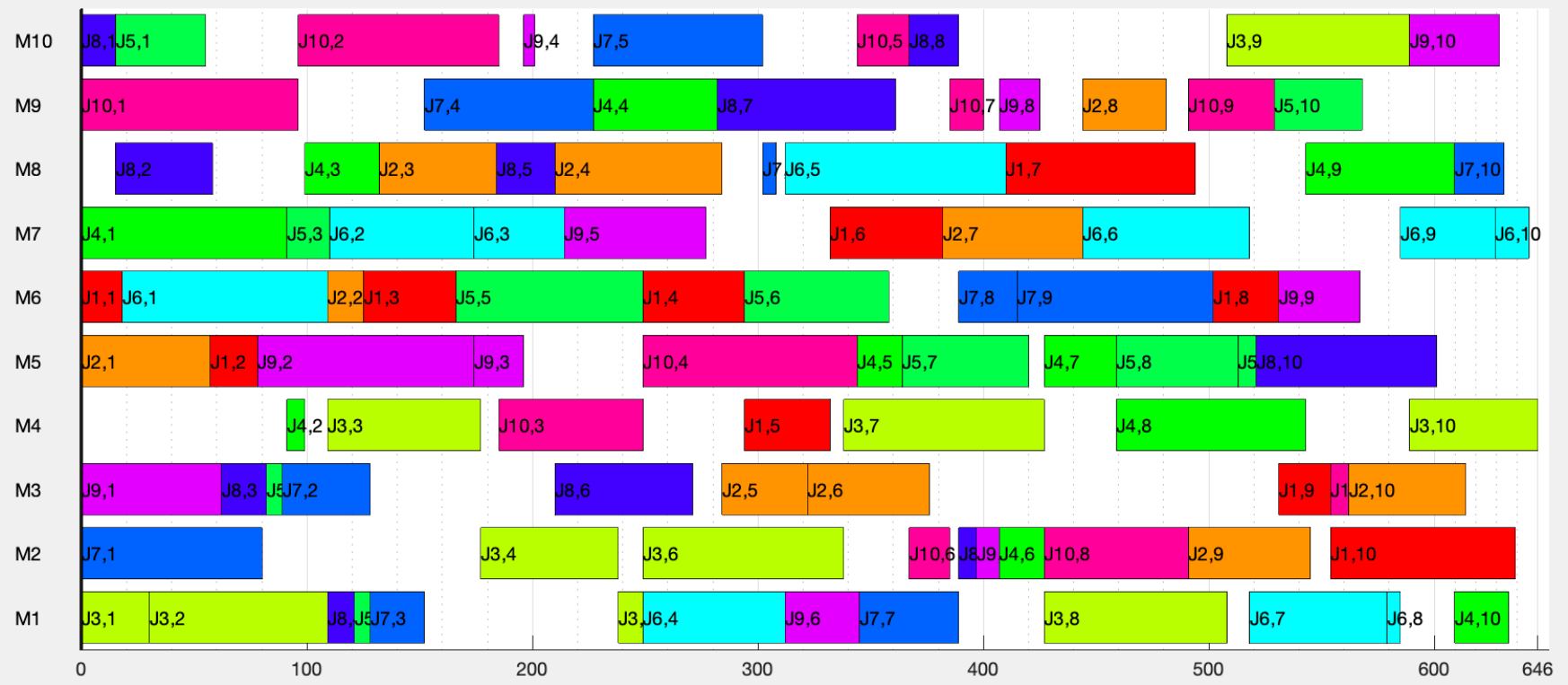}
\caption{Gantt chart of the solution obtained for the instance $la17$-vdata with a makespan of $646$.}
\label{fig:vdata-la17-646}
\end{center}
\end{figure}

\begin{figure}[htbp]
\begin{center}
\includegraphics[scale=0.41]{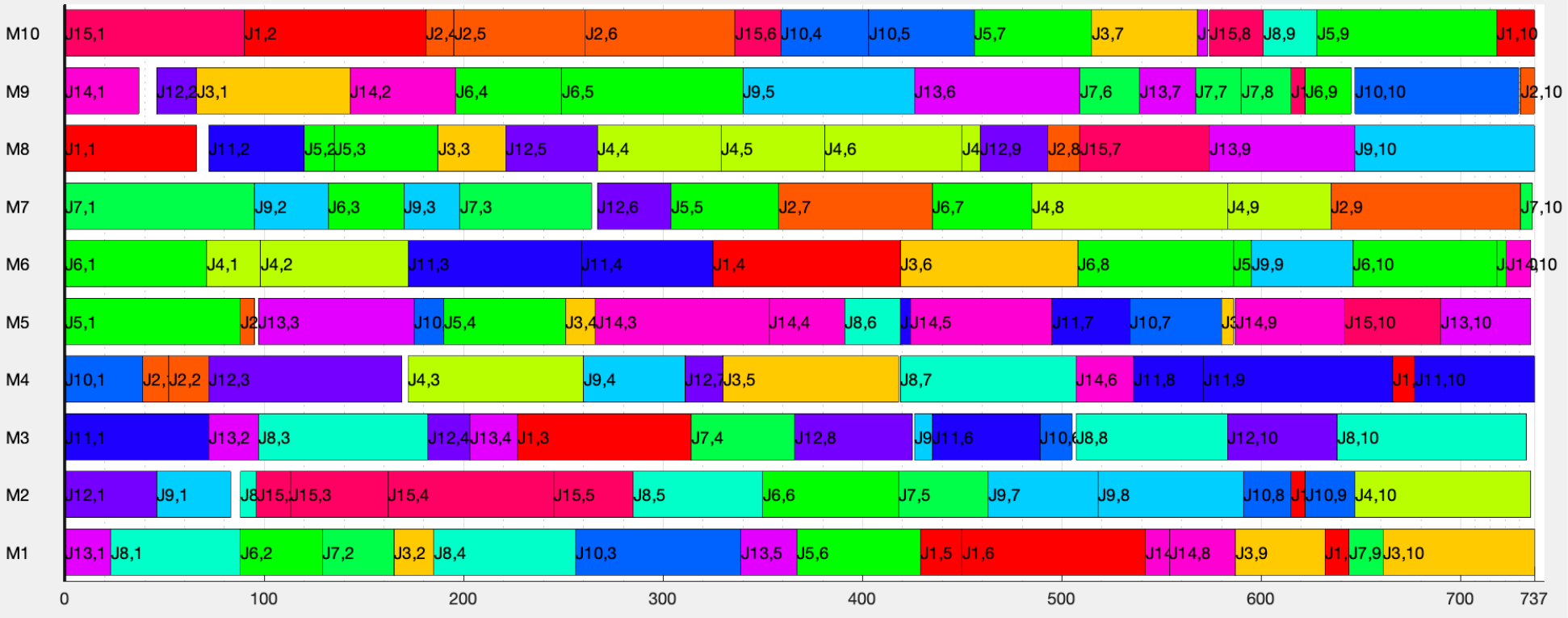}
\caption{Gantt chart of the solution obtained for the instance $la22$-vdata with a makespan of $737$.}
\label{fig:vdata-la22-737}
\end{center}
\end{figure}

\begin{figure}[htbp]
\begin{center}
\includegraphics[scale=0.32]{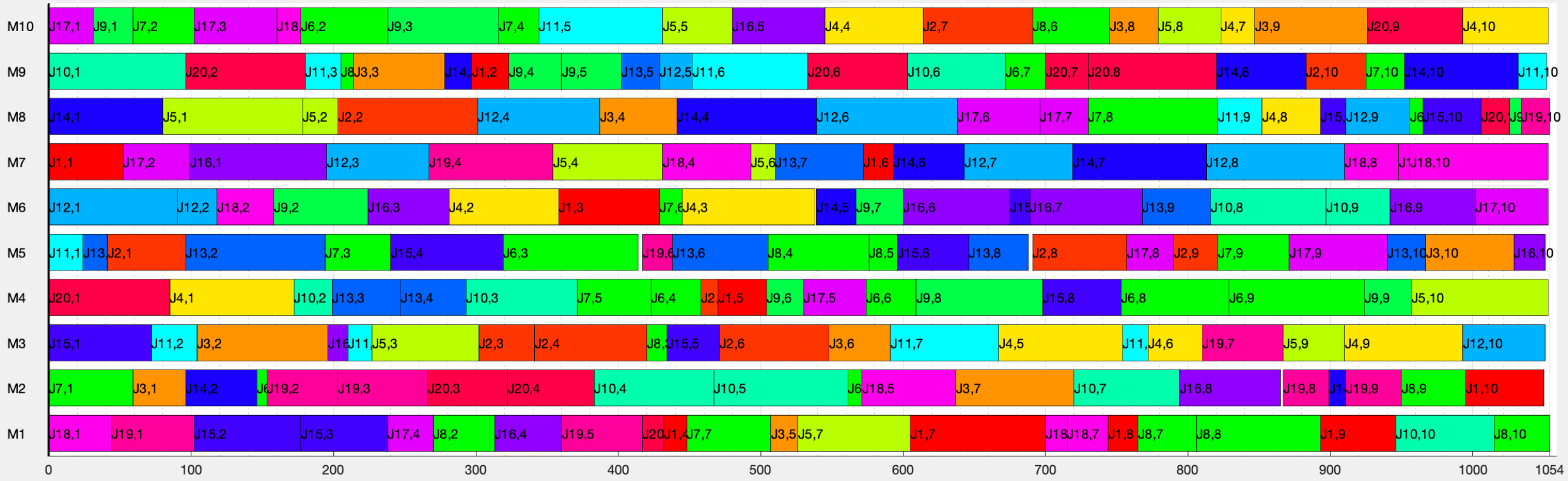}
\caption{Gantt chart of the solution obtained for the instance $la26$-vdata with a makespan of $1054$.}
\label{fig:vdata-la26-1054}
\end{center}
\end{figure}

\begin{figure}[htbp]
\begin{center}
\includegraphics[scale=0.3]{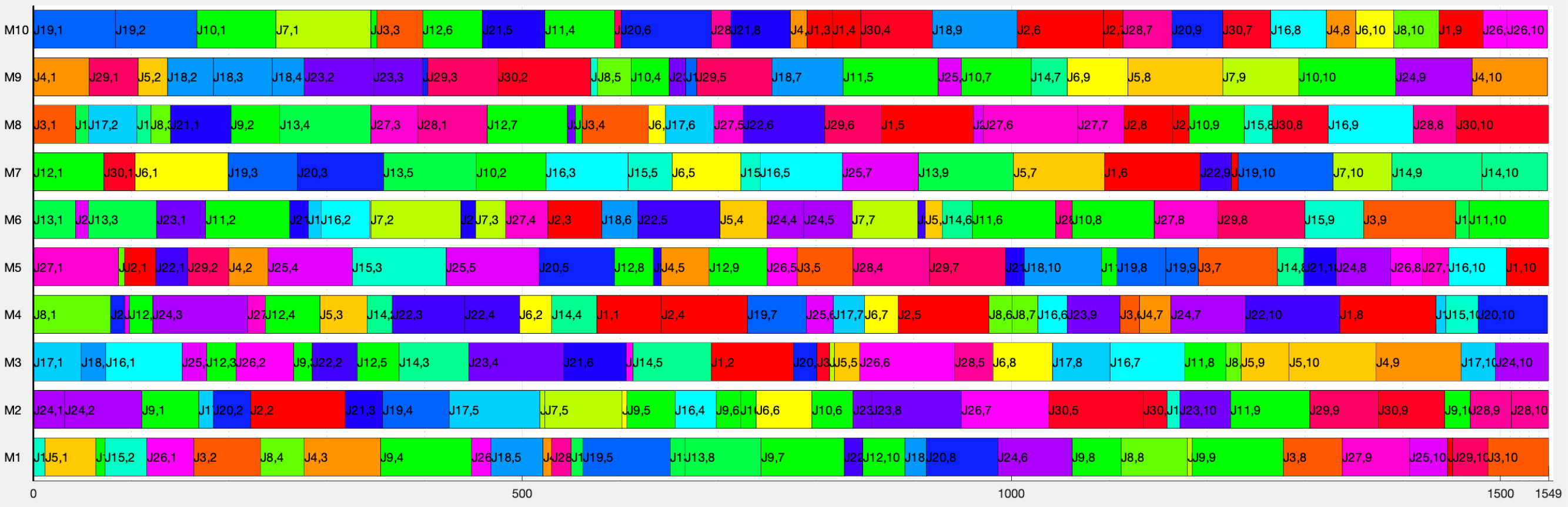}
\caption{Gantt chart of the solution obtained for the instance $la35$-vdata with a makespan of $1549$.}
\label{fig:vdata-la35-1549}
\end{center}
\end{figure}

\begin{figure}[htbp]
\begin{center}
\includegraphics[scale=0.335]{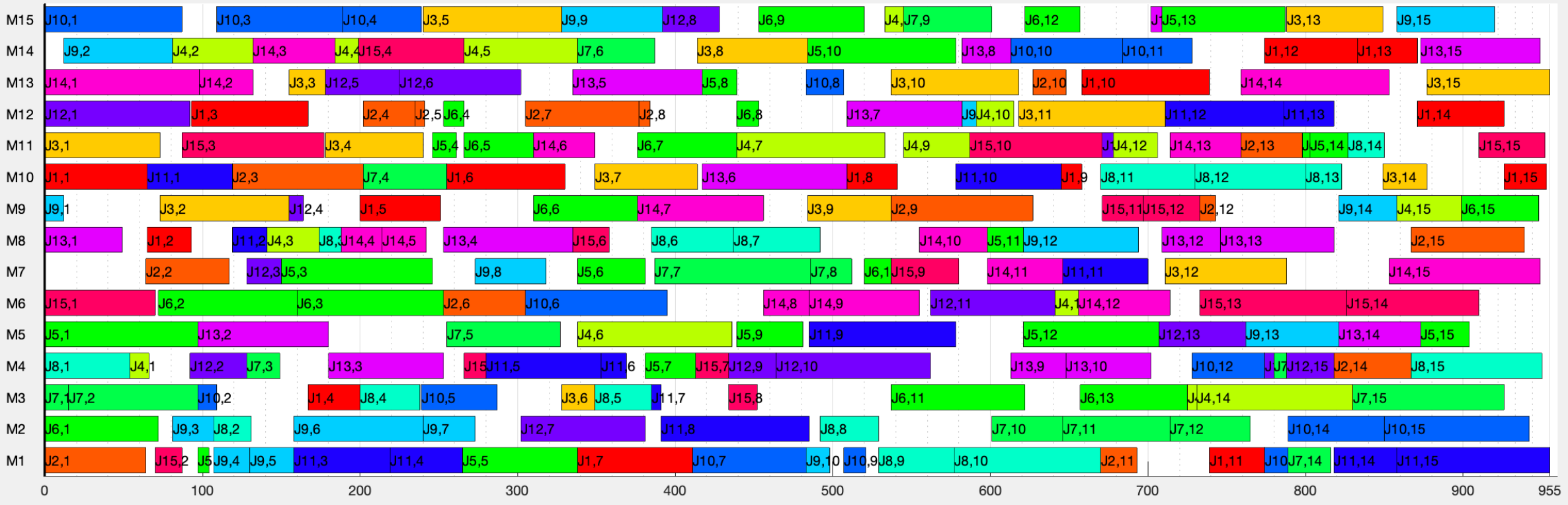}
\caption{Gantt chart of the solution obtained for the instance $la40$-vdata with a makespan of $955$.}
\label{fig:vdata-la40-955}
\end{center}
\end{figure}

\newpage

From the results of both sets of experiments, GLNSA has obtained the best results in most of the problems with a high value of $ \beta $ compared to other algorithms. GLNSA managed to calculate $ 2 $ new makespan-enhancing solutions, particularly for $la15$-rdata and $la22$-vdata, and has the least computational processing time for $ 65$  of $ 86$ instances.

Beyond the different capacities of the computational equipment and the different programming skills and languages used in the implementation of HA (programmed in C ++) and GLNSA (programmed in Matlab and optimized by compiling it with the mex function), the lower computational time of GLNSA compared to that of HA can be explained with the fact that GLNSA performs fewer repetitions of the adaptive way in which the tabu search is applied. While HA handles a value $ T_n = 4$, GLNSA takes $ T_n = 1$, obtaining a lower computational time.

\section{Conclusions and further work}
\label{secc:Conclusions}

This work has presented a hybrid algorithm that performs a global search with $smart\_cells$ using a cellular automaton-like neighborhood where individual operators such as insert and swapping are used, along with an operator like path-relinking, to share the information between solutions . These operators are primarily focused on optimizing the scheduling of operations.

The local search on the GLNSA performs a tabu search to find the best assignment of machines for each operation. Another contribution of this work is that a simplified neighborhood based on Nopt1 is proposed, where the feasible machine of a critical operation is modified without explicitly finding the optimal allocation of the operation, since this is left to global search operations, which is suitable for FJSP instances with high average flexibility $\beta$.

The cellular automaton-like neighborhood allows this type of operations to be carried out concurrently and in a balanced way, which provides a balance between the exploration and exploitation of the GLNSA and allows the use of a lower number of $smart\_cells$ compared to other algorithms, as well as a lower number of iterations of the tabu search, which is reflected in a shorter computational time.

Two well-known benchmarks (including $86$ instances) were used to develop GLNSA's computational experimentation. The results obtained show a good performance and present two best solutions compared to the algorithms taken as a reference.

The GLNSA represents a new way of solving task scheduling, which can be applied to other types of problems, such as the Flowshop, the Job Shop, or the Open Shop Scheduling Problem, where cellular automaton-like neighborhoods can be applied to make concurrent exploration and exploitation actions.

As possible future work, it is proposed to use other operations, such as two-point, POX or JBX crossovers, or other types of mutations, for global search. Other types of local search strategies such as climbing algorithms with restarts can also be used. Also, other simplifications of the Nopt1 neighborhood can be investigated to treat problems with less flexibility.

Finally, the GLNSA approach that uses a cellular automaton-like neighborhood can also be extended to investigate its effectiveness in optimizing multi-objective manufacturing problems.

\section*{Acknowledgement}
This study was supported by the National Council for Science and Technology (CONACYT) with project number CB- 2017-2018-A1-S-43008.

\bibliographystyle{elsarticle-num}
\bibliography{main}

\end{document}